\documentclass{article} 
\usepackage{colm2024_conference}
\usepackage{microtype}
\usepackage{hyperref}
\usepackage{url}
\usepackage{booktabs}
\usepackage{graphicx}
\usepackage{multirow}
\usepackage{xspace}
\usepackage{wrapfig}
\usepackage{amsmath}
\usepackage[most, breakable, many]{tcolorbox}
\usepackage{todonotes}
\usepackage{threeparttable}
\usepackage{indentfirst}

\newcommand{\HippocratesLogo}{\raisebox{-.3em}{\rlap{\raisebox{.3em}{\hspace{1.4em}\scriptsize}}\includegraphics[height=1.8em]
{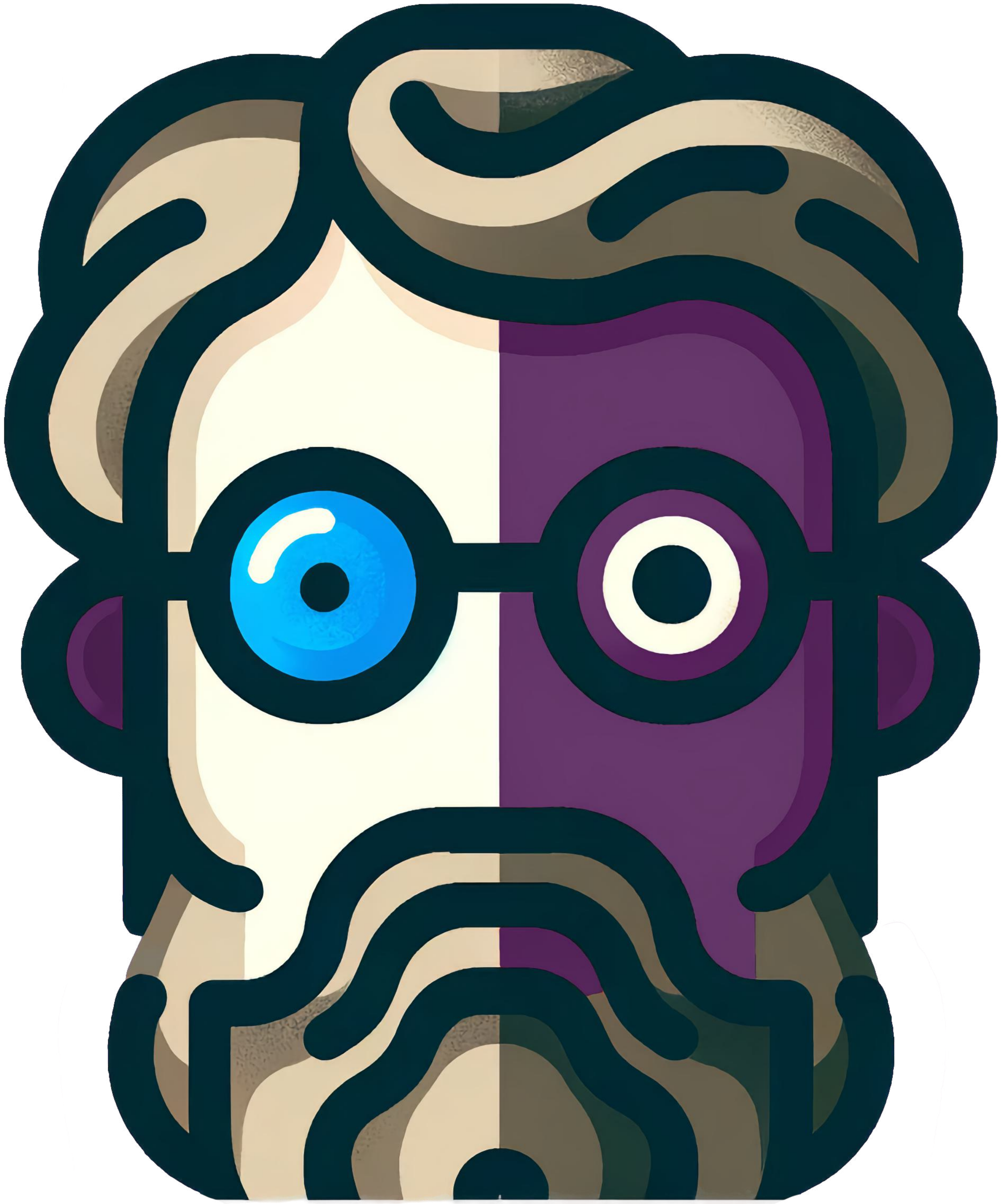}}\xspace}
\newcommand{\LlamaLogo}{\raisebox{-.05em}{\rlap{\raisebox{.05em}{\hspace{1em}\scriptsize}}\includegraphics[height=0.8em]
{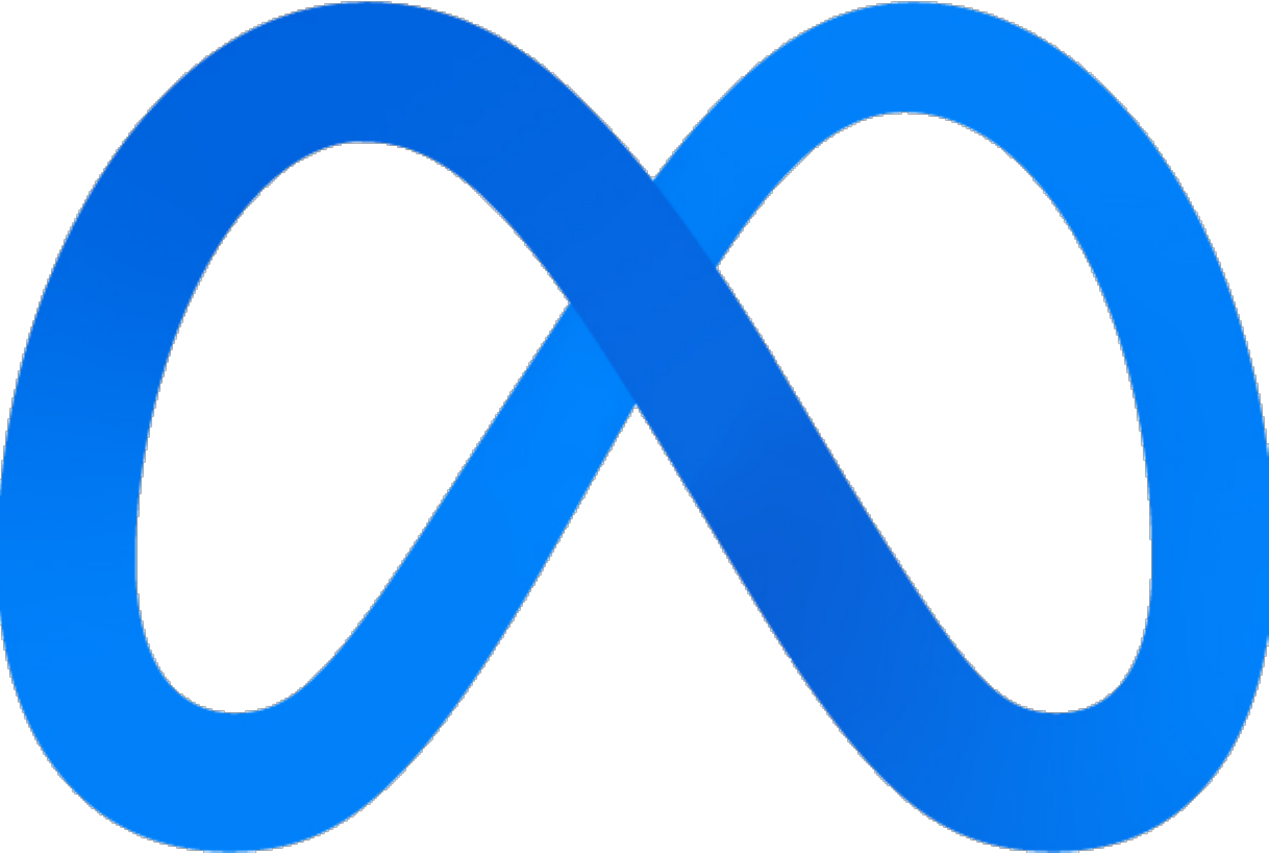}}\xspace}
\newcommand{\MistralLogo}{\raisebox{-.05em}{\rlap{\raisebox{.05em}{\hspace{1em}\scriptsize}}\includegraphics[height=0.8em]
{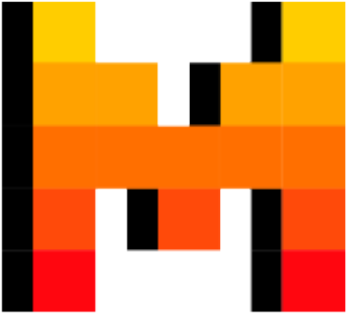}}\xspace}
\newcommand{\HippoL}{\mbox{Hippo-\LlamaLogo}}
\newcommand{\HippoM}{\mbox{Hippo-\MistralLogo}}
\definecolor{mypurple}{HTML}{25004D}
\definecolor{darkgreen}{HTML}{2D8659}
\definecolor{darkred}{HTML}{990000}

\title{\HippocratesLogo~Hippocrates: An Open-Source Framework for Advancing Large Language Models in Healthcare}

\author{Emre Can Acikgoz$^{1,2}$\thanks{Corresponding author, \href{mailto:eacikgoz17@ku.edu.tr}{eacikgoz17@ku.edu.tr}}\;,
    Osman Batur İnce$^{1,2}$\;,
    Rayene Bench$^5$\;,
    Arda Anıl Boz$^4$,
    \textbf{Ilker Kesen}$^1$,\\
    \textbf{Aykut Erdem}$^{1,2}$,
    \textbf{Erkut Erdem}$^{1,3}$ \\[2mm]
    $^1$Koç University, KUIS AI Center, $^2$Koç University, Department of Computer Engineering,\\
    $^3$Hacettepe University, Department of Computer Engineering,  $^4$Robert College\\ $^5$Yıldız Technical University, Department of Computer Engineering \\[2mm]
}

\colmfinalcopy 
\begin{document}

\maketitle

\vspace{-6ex}
\begin{center}
     \url{https://cyberiada.github.io/Hippocrates/}
\end{center}
\vspace{1ex}

\begin{abstract}

The integration of Large Language Models (LLMs) into healthcare promises to transform medical diagnostics, research, and patient care. Yet, the progression of medical LLMs faces obstacles such as complex training requirements, rigorous evaluation demands, and the dominance of proprietary models that restrict academic exploration. Transparent, comprehensive access to LLM resources is essential for advancing the field, fostering reproducibility, and encouraging innovation in healthcare AI. We present Hippocrates, an open-source LLM framework specifically developed for the medical domain. In stark contrast to previous efforts, it offers unrestricted access to its training datasets, codebase, checkpoints, and evaluation protocols. This open approach is designed to stimulate collaborative research, allowing the community to build upon, refine, and rigorously evaluate medical LLMs within a transparent ecosystem. Also, we introduce Hippo, a family of 7B models tailored for the medical domain, fine-tuned from Mistral and LLaMA2 through continual pre-training, instruction tuning, and reinforcement learning from human and AI feedback. Our models outperform existing open medical LLMs models by a large-margin, even surpassing models with 70B parameters. Through Hippocrates, we aspire to unlock the full potential of LLMs not just to advance medical knowledge and patient care but also to democratize the benefits of AI research in healthcare, making them available across the globe. 

\end{abstract}


\section{Introduction}
\label{sec:introduction}


The remarkable success of Large Language Models (LLMs) across diverse NLP tasks has revolutionized artificial intelligence \citep{llama22023, Bai2023QwenTR, mistral7b, gpt4, gemini}. Despite their impressive generalization capabilities, LLMs encounter challenges in clinical contexts, primarily due to a deficiency in domain-specific knowledge and the intricacies of medical terminology. Bridging this gap, in this work, we introduce Hippocrates (named after the Ancient Greek ``Father of Medicine"), a state-of-the-art, fully open-source framework designed to elevate LLMs' proficiency in medical reasoning. We publicly share our training data, complete training and evaluations codes, along with intermediate model checkpoints. Our framework marks an important step towards democratizing advancements in medical LLMs.

Previous attempts to develop advanced medical LLMs yielded promising results by further training them \citep{biomistral}, supervised fine-tuning them \citep{chatdoctor2023, medalpaca2023, clinicalcamel2023}, or both \citep{pmcllama2023, meditron70b}, via special medical-text corpus and medical instruction datasets. However, the data collection, pre-training, and finetuning stages may include considerable complexity, which makes reproducing, analyzing, and comparing the recent LLMs in that domain challenging. On the other hand, closed models, e.g. GPT4~\citep{gpt4}, Gemini~\citep{gemini}, Med-PaLM~\citep{medpalm2023}, trained on closed-domain datasets make their results non-reproducible, not to mention substantial computational costs and further complicate the understanding of which components are crucial to the success of these advanced medical frameworks.

\begin{figure}[!t]
\centering
\includegraphics[width=\linewidth]{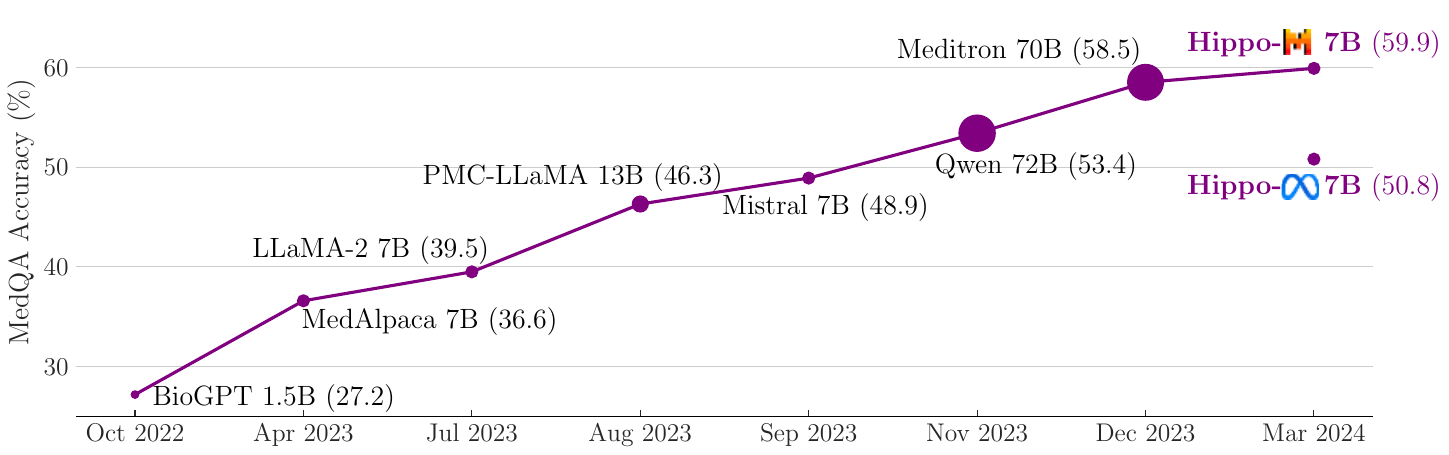}
\caption{\textbf{The evolution of medical LLM performances on the MedQA dataset.} Our 7B \HippoL~and \HippoM~models achieve 50.8\% and 59.9\% 5-shot accuracy, respectively. \HippoM~outperforms all existing open models, including even those with 70B parameters.}
\label{fig:progress}
\end{figure}


In this work, we provide full access to our framework, from the data sources to the training configurations and the reproducible evaluation protocols. We conduct a detailed empirical analysis to identify the impact of various design elements on LLM performance, leading to a domain-adapted framework that demonstrates superior performance on multiple medical benchmarks. Based on these insights, we develop a step-by-step guide for the efficient training of medical-LLMs. Our research efforts yield two advanced 7B parameter models, \HippoL~and \HippoM. As shown in Fig.~\ref{fig:progress}, our models not only outperform existing 7B and 13B models by a significant margin but also deliver results on par with, and in some cases exceeding, those of 70B models.
We argue that the development of a broad, varied collection of open models is crucial for deepening our knowledge of language models and enhancing their applicability across various domains.

In addition, we adopt a novel strategy for structuring our instruction tuning (IT) dataset, dividing it into two distinct components: the General Instruction Dataset and the Evaluation Instruction Dataset. The General dataset is designed to enable unbiased assessments by avoiding overlap with downstream task data, marking a departure from previous methodologies. On the other hand, the Evaluation Instruction Dataset, which incorporates training splits from evaluation benchmarks, facilitates direct comparisons with existing models \citep{meditron70b}. Notably, for the first time in the medical domain, our approach incorporates preference learning from medical professionals into the model development process, utilizing RLAIF \citep{rlaif} and GPT4 for annotating preferences.

For model evaluation, we employ the well-established EleutherAI framework\footnote{\url{https://github.com/EleutherAI/lm-evaluation-harness}} \citep{eval-harness}, conducting tests across a set of six varied medical downstream tasks. These include MedMCQA \citep{medmcqa}, PubmedQA \citep{pubmedqa}, MedQA \citep{medqa}, and the USMLE-step1, USMLE-step2, and USMLE-step3. Leveraging this framework allows for straightforward replication of any LLM's results, eliminating the necessity for additional fine-tuning or the repetitive execution of evaluation scripts for each new model.

\section{Hippocrates Framework}
\label{sec: framework}
\begin{figure}[!t]
\centering
\includegraphics[width=\linewidth]{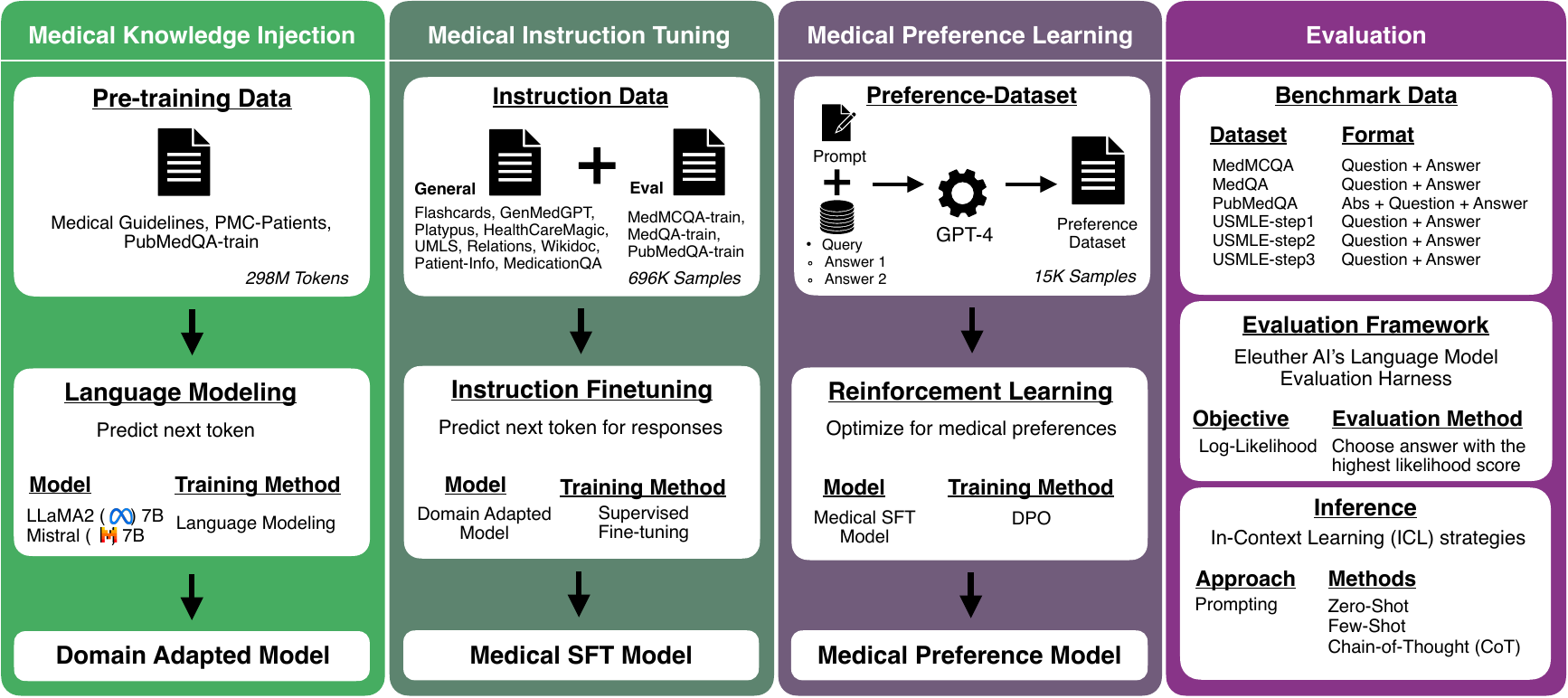}
\caption{\textbf{An overview of the Hippocrates framework}, illustrating the four critical phases including (1) continued pre-training, (2) supervised fine-tuning, (3) reinforcement learning from AI-generated feedback, and (4) the comprehensive evaluation pipeline.}
\label{fig:overview}
\end{figure}

Fig.~\ref{fig:overview} shows the overall workflow of the Hippocrates framework, starting from domain-specific pre-training and progressing through supervised fine-tuning and reinforcement learning from AI-generated feedback to an extensive evaluation phase. This pipeline ensures our models are precisely tailored and rigorously tested for the medical domain.

\subsection{Continued Pre-training Data}
\label{sec: framework-pre-training-data}

A key aspect of our methodology is the integration of specialized medical knowledge through an extensive pre-training corpus, assembled from three specialized datasets: Medical Guidelines, PMC-Patients, and PubMedQA-contexts. The Medical Guidelines dataset comprises clinical practice guidelines, is used for training Meditron models \citep{meditron70b}. The PMC-Patients dataset \citep{pmcpatients-dataset} consists of patient summaries extracted from case reports within PubMed Central (PMC). Additionally, the PubMedQA-contexts dataset is constructed by extracting the context field of each sample in the training split of the benchmark \citep{pubmedqa}. Detailed descriptions and specifications of each dataset are available in Table \ref{tab:dataset-pre-training}. This extensive corpus, consisting of roughly 300M training tokens, forms the foundation of our models, ensuring their proficiency in navigating medical terminology and practices. We systematically assessed the impact of each dataset, both individually and in combination, to optimize our model's performance.

\renewcommand{\arraystretch}{1.15}

\begin{table}[h]
\centering
\resizebox{\linewidth}{!}{
\begin{tabular}{@{$\;$}ll@{$\;\;$}l@{$\;\;$}r@{$\;\;$}r@{$\;\;$}r@{$\;$}}
\hline
\textbf{Dataset}     & \textbf{Source}       & \textbf{License} & \textbf{Size (MB)}        & \textbf{\#Samples} & \textbf{\#Tokens}         \\ \toprule
Medical Guidelines   & Meditron              & Apache 2.0 License   & 382.6                     & 37,970                 & 96M                        \\
PMC-Patients         & Pubmed Central        & CC BY-NC-SA 4.0 & 462.3                     & 167,034                & 122M                            \\
PubMedQA-train       & PubMedQA              & MIT License & 290.2                     & 211,269                & 80M                                 \\ \midrule
\textbf{Total} & & & \textbf{1,135.1} & \textbf{416,273} & \textbf{298M} \\
\bottomrule
\end{tabular}%
}
\caption{\textbf{Summary of the datasets used for continued pre-training}, showing their sources, licence information and data statistics.}
\label{tab:dataset-pre-training}
\end{table}

\subsection{Supervised Fine-Tuning Data}
\label{sec: framework-instruction-data}
Developing effective medical LLMs requires blending domain-specific knowledge with sophisticated reasoning abilities. Previous models often utilized instruction data consisting of samples from the training or test sets of evaluation benchmarks. We also considered this setup, but additionally investigated an alternative involving generic medical data. Consequently, we constructed two sets of IT datasets: the General Instructions Data and the Evaluation Instructions Data.

\paragraph{General Instructions Data.} This dataset aggregates more than 400K samples from nine different datasets, each derived from the instruction corpora of previous studies \citep{chatdoctor2023, medalpaca2023, pmcllama2023, platypus2023}. By excluding data from the training or test splits of downstream QA benchmarks, we aim to minimize bias and improve the model's generalization capabilities across different reasoning tasks. A pre-processing protocol was employed to remove 
superfluous words and web URLs, ensuring the data's quality and relevance. The detailed statistics of the dataset are presented in Table \ref{tab: dataset-instruction}.

\begin{table}[h!]
\resizebox{\linewidth}{!}{
\begin{tabular}{@{$\;$}ll@{$\;\;$}l@{$\;\;$}r@{$\;\;$}r@{$\;\;$}r@{$\;$}}
\hline
\textbf{Dataset}     & \textbf{Source}       & \textbf{License} & \textbf{Size (MB)} & \textbf{\#Samples}        & \textbf{\#Tokens}             \\ \toprule
Medical Flashcards   & MedAlpaca             & No commercialized use & 18.8                      & 33,955                 & 3.9M                         \\
GenMedGPT-5k         & ChatDoctor             & Apache 2.0    & 3.1                       & 5,452                  & 0.6M                               \\
Open-Platypus        & Platypus              & CC BY-NC-SA 4.0  & 32.9                      & 24,926                 & 9.5M                             \\
HealthCareMagic-100k & ChatDoctor            & Apache 2.0  & 143.8                     & 112,165                & 32.3M                                 \\
UMLS                 & PMC-LLaMA             & CC BY 4.0 & 23.0                      & 49,057                 & 4.6M                                          \\
UMLS-Relations       & PMC-LLaMA             & CC BY 4.0 & 21.7                      & 50,000                 & 4.3M                                          \\
WikiDoc              & MedAlpaca             & CC BY-SA 4.0 & 11.0                      & 10,000                 & 2.6M                                  \\
WikiDoc-Patient-Info & MedAlpaca             & CC BY-SA 4.0  & 3.7                       & 5,942                  & 0.8M                                 \\
MedicationQA         & PMC-LLaMA            & CC BY 4.0 & 0.4                       & 552                    & 0.1M   \\ \midrule
\textbf{Total} & & & \textbf{258.4} & \textbf{292,049} & \textbf{58.7M}\\ \bottomrule
\end{tabular}%
}
\caption{\textbf{Summary of General Instructions Data}, describing the datasets used, their sources, together with their licence information, and size.}
\label{tab: dataset-instruction}
\end{table}


\paragraph{Evaluation Instructions Data.} This dataset was formed to examine the effects of including instruction samples directly from downstream tasks, a common practice in existing studies \citep{meditron70b, medalpaca2023, pmcllama2023}. Instruction-response pairs were crafted using the training splits of various benchmarks, following the templates established in Meditron~\citep{meditron70b}. We conducted a series of experiments to assess the distinct influence of each split on each task, both individually and collectively. The details about the Evaluation Instruction Data is given in Table \ref{tab: dataset-instruction-eval}.

\begin{table}[h!]
\centering
\resizebox{0.95\linewidth}{!}{
\begin{tabular}{lllrrr}
\hline
\textbf{Dataset}     & \textbf{Source}      & \textbf{License}  & \textbf{Size (MB)}   & \textbf{\#Samples}   & \textbf{\#Tokens}       \\ \toprule
MedMCQA-train        & MedMCQA               & MIT License    & 114.4                & 182,822                  & 24.9M                         \\
MedQA-train          & MedQA                  & MIT License  & 14.2                 & 10,178                   & 3.4M                           \\
PubMedQA-train       & PubMedQA               & MIT License & 76.3                 & 211,269                  & 95.9M                            \\ \bottomrule
\textbf{Total}       &  &  & \textbf{204.9}                & \textbf{404,269} & \textbf{124.2M}                                 \\ \bottomrule
\end{tabular}%
}
\caption{\textbf{Summary of Evaluation Instructions dataset}, showing which training splits of the downstream tasks they are derived from and their data statistics.}
\label{tab: dataset-instruction-eval}
\end{table}
Beyond independently utilizing these datasets for supervised fine-tuning, we also examined the impact of individual datasets as well as the collective effect of combining them on model performance (refer to Appendix~\ref{sec: appendix-training-stages}).



\subsection{Medical Preference Data}
\label{sec: medical-comparison-data}
Constructing a preference dataset typically involves generating diverse responses to identical queries using LLMs, which are subsequently evaluated by human annotators to identify the most accurate response. This method, however, can become prohibitively expensive, both in terms of computation for generating responses and the financial and time investments required for manual annotation. To circumvent these issues, we leveraged the iCliniq-10k dataset \citep{chatdoctor2023}, containing 10K authentic patient-doctor dialogues from icliniq.com. Each dialogue features a patient question accompanied by three different answers: one from an actual doctor, and the others from ChatGPT and ChatDoctor~\citep{chatdoctor2023}. We conducted a thorough preprocessing of this dataset to eliminate any irrelevant or extraneous information.

\paragraph{Medical RLAIF.} To reduce annotation costs, we adopted the RLAIF methodology \citep{rlaif} in the medical domain for the first time. Utilizing detailed prompts based on patient inquiries from the iCliniq-10k dataset, we used GPT4~\citep{gpt4} to determine the optimal response based on predefined instructions. These instructions were derived from those used in qualitative assessments by medical professionals in Med-PaLM~\citep{medpalm2022, medpalm22023}, with minor modifications. This annotation approach amounted to a cost of \$120. The exact prompt structure for applying RLAIF with GPT4 is given in Appendix~\ref{sec: negative-results}, Figure \ref{fig:RLHAIF-prompt}.

\paragraph{Validation.} To test the reliability of GPT4's capacity to replicate medical expert annotations, we subjected 250 samples from our dataset to careful examination by two medical doctors, given them the same instructions that we provided in the prompt to GPT4.
Our analysis revealed compelling results. When comparing GPT4's annotations against those of MD-1, GPT4 demonstrated a Kappa Score of 0.376, indicating moderate agreement, and an accuracy of 68.9\%. The comparison with MD-2 showed even stronger results, with GPT4 achieving a Kappa Score of 0.672, suggesting substantial agreement, alongside an 83.6\% accuracy. Interestingly, the inter-annotator agreement between the two doctors themselves yielded a Kappa Score of 0.416 and an accuracy of 70.8\%, situating GPT4's performance firmly within the range of human expert variability. These findings not only affirm GPT4's aptitude for medical annotation but also highlight its potential to serve as a cost-effective alternative to human annotators in medical research and application settings. These findings suggest that GPT4 is capable of effectively mimicking medical doctor preferences, potentially eliminating the need for costly doctor annotations. 

Consequently, we compiled a comprehensive medical doctor preference dataset, consisting of 15,258 samples, to further align our LLMs with real-world clinical decision-making processes and enhance their accuracy in interpreting and responding to medical queries.



\subsection{Training Methodology}
\label{sec: framework-model-training}
Our training strategy includes several phases: injection of medical knowledge through continued pre-training, domain-specific instruction tuning, and reinforcement learning from AI-generated feedback for improved alignment with medical experts. Employing the LLaMA Factory framework~\citep{llama-factory}, we adhere to replicable and high-performance training standards. Moreover, we adopt the Low-Rank Adaptation (LoRA) technique~\cite{lora2021} for training efficiency and precision. LoRA enhances LLMs by selectively updating weights within additional trainable layers, thereby accelerating the training process, minimizing memory usage, and mitigating overfitting and catastrophic forgetting.

Our foundational models, LLaMA2 7B~\citep{llama22023} and Mistral 7B~\citep{mistral7b}, are selected based on their robust performance across medical benchmarks, demonstrating their capacity to excel without extensive training modifications. The zero-shot performances of these generic baseline models is presented at the beginning of Table \ref{tab: main-results-zero-shot}.

\paragraph{Continued pre-training.} To equip our base LLMs with domain-specific medical expertise, we extend their pre-training on a carefully curated medical text corpus as described in Section \ref{sec: framework-pre-training-data}. This stage employs traditional language modeling, focusing on next-token prediction. During this phase, both models undergo continued pre-training using LoRA, specifically adapting the fully connected layers. The parameters for LoRA are carefully set, with the rank ($r$) at 8 and alpha ($\alpha$) at 16, to optimize learning. We use the AdamW optimizer and adjust the learning rate using a cosine scheduling, starting from an initial value of 1e-4. The batch size per device was initialized to be 8, with gradient accumulations of 2, culminating in an effective global batch size of 16, and the models are trained for a single epoch. The rationale and empirical support for our choices regarding the dataset, LoRA configurations, and overall optimization strategy are comprehensively analyzed in Appendix \ref{sec: appendix-continued-pre-training}.

\paragraph{Supervised Finetuning.} After continued pre-training, models undergo fine-tuning with an Instruction Tuning (IT) dataset to closely mirror medical directives, aligning model outputs with clinical requirements. We have tested with the datasets described in Section~\ref{sec: framework-instruction-data} and found that MedQA-train IT works better than the other options. This fine-tuning phase also employs LoRA to all fully connected layers with both rank ($r$) and alpha ($\alpha$) set to 32 for balanced efficiency and computational overhead. AdamW optimizer is used with a learning rate of $1e-4$. To prevent model overfitting, loss calculation focuses solely on the responses. The training spanned 3 epochs with a batch size of 8 per-device and gradient accumulation set to 2. We also conducted experiments on direct fine-tuning of the base LLMs to evaluate the impact of continued pre-training (see Section \ref{sec: contribution-of-each-stage}) and performed a comprehensive analysis on dataset splits and fine-tuning hyperparameters (see Appendix~\ref{sec: appendix-sft}).

\paragraph{Medical Preference Learning.} Finally, the instruction-tuned models are further trained with a recent and popular technique called direct preference optimization (DPO) \citep{dpo}. In DPO, reinforcement learning is bypassed which allows for direct optimization based on preference data. Unlike RLHF, the responses in DPO need not be derived from the LLM being optimized. Central to DPO is the development of a loss function that evaluates the likelihood of a preferred response over a less preferred one, steering the LLM towards this goal. This makes DPO more stable and significantly reduces computational demands. 

The outcome of all this are our medical LLMs, named \HippoL~and \HippoM, built upon the pre-trained LLaMA2 7B and Mistral 7B models. These models were refined through a comprehensive process that included continued pre-training and/or instruction tuning using our carefully curated medical datasets. Following this, we also explored the impact of aligning the models with clinical preferences by conducting further training on medical preference data.

\section{Main Results}
\label{sec: main-results}
For an objective evaluation of domain-specific knowledge and reasoning capabilities in LLMs, a detailed and fair evaluation framework is essential. In alignment with methodologies adopted in prior research \citep{medpalm2022, medalpaca2023, pmcllama2023, clinicalcamel2023, medpalm22023, meditron70b}, we selected six widely recognized medical question-answering datasets, namely MedMCQA \citep{medmcqa}, MedQA \citep{medqa}, PubMedQA \citep{pubmedqa} and USMLE Step 1-3 \citep{medalpaca2023}, to assess models performances (See Table~\ref{tab: dataset-evaluation} for details). Performance metrics were derived through the use of the EleutherAI evaluation framework \citep{eval-harness}, ensuring a standardized approach to measuring model effectiveness in handling domain-specific queries.

\begin{table}[h!]
\resizebox{\linewidth}{!}{
\begin{tabular}{lllrrr}
\hline
\textbf{Dataset}     & \textbf{Source}       & \textbf{Format}                & \textbf{\#Samples}  & \textbf{\#Choices} & \textbf{License}      \\ \toprule
MedMCQA-test         & MedMCQA               & Question + Answer              & 4,183             & 4                        & MIT            \\
MedQA-test           & MedQA                 & Question + Answer              & 1,273             & 5                         & MIT            \\
PubMedQA-test        & PubMedQA              & Abstract + Question + Answer   & 1,000             & 3                       & MIT            \\
USMLE-step1          & USMLE                 & Question + Answer              & 94                & 5                         & MIT            \\
USMLE-step2          & USMLE                 & Question + Answer              & 109               & 6                         & MIT            \\
USMLE-step3          & USMLE                 & Question + Answer              & 122               & 5                      & MIT            \\ \bottomrule
\end{tabular}%
}
\caption{\textbf{Summary of the evaluation benchmark datasets}, describing the format, the number of test samples, the number of choices, and the licence info.}
\label{tab: dataset-evaluation}
\end{table}


\vspace{-2ex}

\subsection{Experimental Setup}
\label{sec: experimental-setup}
In our evaluation, we included a spectrum of leading LLMs, spanning general and medical LLMs, varying in scale from 1.5B to an advanced 70B parameters. Here we report the performances of our top-performing models for an accurate comparison. To ensure a fair and easily replicable assessment of these medical models, we utilized the Eleuther AI Language Model Evaluation Harness \citep{eval-harness}, a unified evaluation framework specifically designed for evaluating generative LLMs. This framework also serves as the evaluation tool for the Open LLM Leaderboard\footnote{\url{https://huggingface.co/spaces/HuggingFaceH4/open_llm_leaderboard}} \citep{open-llm-leaderboard}.

LM-Evaluation-Harness operates on a Log-Likelihood objective, which calculates the negative log-likelihood for each potential answer in response to a given query. The answer is then chosen based on the highest likelihood score, indicating it as the most probable choice. During evaluation, each prompt includes a question and corresponding choices, separated by a new line. For PubMedQA, the abstract provides contextual grounding for the model's decision-making process. Examples of these prompts are provided in the Appendix~\ref{sec:appendix-eval}. 

\subsection{Results}
\begin{table}[t]
\resizebox{\linewidth}{!}{
\begin{tabular}{@{}l@{$\;\;$}c@{$\;\;$}c@{$\;\;$}c@{$\;\;$}c@{$\;\;$}c@{$\;\;$}c@{$\;\;$}c@{}}
\toprule
\multirow{2}{*}{\textbf{Model}} & \textbf{MedMCQA} & \textbf{MedQA} & \textbf{PubmedQA} & \textbf{USMLE-1} & \textbf{USMLE-2} & \textbf{USMLE-3} &\textbf{Avg.} \\ 
& 0-shot/5-shot & 0-shot/5-shot & 0-shot/5-shot & 0-shot/5-shot & 0-shot/5-shot & 0-shot/5-shot & 0-shot/5-shot\\
\toprule
Gemma 2b          & 26.2/27.7   & 27.8/30.6   & 59.1/60.8                & 20.2/16.0                 & 18.4/30.3                         & 24.6/20.5               & 29.4/31.0         \\
LLaMA-2 7b        & 34.4/39.4   & 29.3/39.5   & 72.3/72.4                & 18.1/22.3                 & 22.9/33.0                         & 27.1/32.0               & 34.0/39.8         \\
Falcon 7b         & 30.5/31.8   & 27.9/31.0   & 65.3/64.4                & 18.1/25.5                 & 26.6/20.2                         & 23.8/25.4               & 32.0/33.0         \\
Vicuna 7b         & 35.9/39.0   & 35.1/41.2   & 70.9/74.5                & 25.5/31.9                 & 27.5/31.2                         & 33.6/35.3               & 38.1/42.2        \\
Mistral 7b        & 39.3/48.5   & 36.8/48.9   & 76.3/77.8                & 24.5/50.0                 & 31.2/42.2                         & 27.9/43.4               & 39.3/51.8       \\  
BioMedLM          & 32.2/29.6   & 29.3/30.6   & 55.2/55.2                & 15.9/22.3                 & 19.3/18.4                         & 23.0/31.2               & 25.9/31.2         \\
BioGPT-Large      & 33.1/30.1   & 31.3/27.2   & 60.1/47.7                & 22.3/19.2                 & 22.0/14.7                         & 23.0/23.0               & 32.0/27.0         \\
MedAlpaca 7b      & 35.8/37.5   & 36.1/36.6   & 73.2/70.6                & 22.3/27.7                 & 27.5/32.1                         & 29.5/37.7               & 37.4/40.4         \\
PMC-LLaMA 7b      & 31.5/33.0   & 28.0/29.5   & 66.5/68.4                & 21.3/19.2                 & 23.9/19.3                         & 22.1/22.1               & 32.2/31.9         \\
Meditron 7b       & 34.0/38.2   & 32.0/39.3   & 71.6/75.7                & 16.0/29.8                 & 25.7/30.3                         & 23.8/32.0               & 33.9/40.9         \\
Bio-Mistral 7b    & 36.4/42.4   & 35.0/42.1   & 73.4/75.1                & 24.5/28.7                 & 27.5/34.9                         & 27.9/44.3               & 37.5/31.9 \\ \midrule
LLaMA-2 13b       & 38.2/43.9   & 34.3/43.3   & 75.9/71.9                & 20.2/38.3                 & 22.0/29.4                         & 23.0/38.5               & 35.6/40.9         \\
Vicuna 13b        & 39.7/44.3   & 35.9/45.9   & 75.6/75.0                & 24.5/40.4                 & 26.6/35.8                         & 23.8/46.7               & 37.7/44.6       \\ 
MedAlpaca 13b     & 32.5/33.3   & 31.8/34.3   & 72.6/72.5                & 24.5/23.4                 & 24.5/26.6                         & 30.3/29.5               & 36.0/44.2         \\
PMC-LLaMA 13b     & 39.1/44.5   & 37.8/46.3   & 76.8/76.5   & 30.9/35.1                 & 22.9/36.7                         & 26.2/29.5               & 39.0/44.8 \\  \midrule
LLaMA-2 70b       & 42.8/ 52.0  & 44.9/56.1   & 73.2/77.8                & 31.9/59.6                 & 44.0/57.8                         & 44.3/53.3               & 46.8/59.4      \\
Qwen 72b          & \underline{50.5}/\textbf{59.2}  & 47.7/53.4   & \underline{77.2}/76.8   & 45.7/\textbf{67.0}                 & \underline{43.1}/56.9                         & 38.5/\textbf{61.5}     & 50.5/\textbf{62.5}      \\
ClinicalCamel 70b & 43.7/53.4   & 45.5/\underline{58.5}   & 73.6/77.6                & 40.4/59.6     & \underline{43.1}/\underline{60.6} & 42.6/\underline{60.7}    & 48.2/61.7         \\
Meditron 70b      & 43.4/51.9   & 44.9/\underline{58.5}   & 76.4/\textbf{80.0} & 35.1/57.5 & 41.3/56.9                         & 37.7/59.8   & 46.5/60.8 \\ \midrule
\HippoL~7b        & \textbf{54.3}/\underline{53.9} & \underline{50.6}/50.8    & 74.7/76.6 & \underline{46.8}/40.4  & 41.3/39.5   & \underline{50.0}/43.4   & \underline{53.0}/50.8     \\
\HippoM~7b        & 49.7/51.8  &\textbf{59.2}/\textbf{59.9}& \underline{77.1}/\underline{78.1} & \textbf{60.6}/\underline{61.7}& \textbf{66.1}/\textbf{64.2}& \textbf{56.6}/56.6& \textbf{61.6}/\underline{62.1} \\ \bottomrule
\end{tabular}
}
\caption{\textbf{Comparative analysis of generic and medical LLMs across downstream medical tasks in 0-shot and 5-shot learning settings.} The best and the second-best performance are highlighted in bold and underline, respectively.}
\label{tab: main-results-zero-shot}
\end{table}

We present a comparative analysis of our novel models, \HippoL~and \HippoM, against a set of established base LLMs and medical-specific LLMs, in Table~\ref{tab: main-results-zero-shot}. Our evaluation includes both zero-shot and few-shot (specifically, 5-shot) learning scenarios. Demonstrating superior performance, our Hippo models outperform traditional pretrained models in zero-shot evaluations and maintain their superiority in the 5-shot context. Remarkably, \HippoL~and \HippoM~not only beat models with 7 billion and 13 billion parameters but also exceed the capabilities of those with 70 billion parameters. This outstanding performance highlights the adaptability and precision of our models, showing their remarkable ability to significantly boost prediction accuracy with minimal input examples.



\section{Analysis}
\label{sec: analysis}

\subsection{Contribution of Each Training Stage}
\label{sec: contribution-of-each-stage}

\begin{table*}[!t]
\resizebox{\linewidth}{!}{
\begin{tabular}{lc@{$\;\,$}c@{$\;\,$}c@{$\;\,$}c@{$\;\,$}c@{$\;\,$}c@{$\quad$}c}
\toprule
\textbf{Model}                 & \textbf{MedMCQA}  & \textbf{MedQA}  & \textbf{PubmedQA}  & \textbf{USMLE-1}  & \textbf{USMLE-2}  & \textbf{USMLE-3}  &\textbf{Avg.}  \\ \midrule
LLaMA2 7b                      & 34.4              & 29.3            & 72.3               & 18.1              & 22.9              & 27.1              & 34.0          \\  
$\;\;$ + CP                    & 34.6              & 31.9            & 72.8               & 20.2              & 25.7              & 21.3              & 34.4          \\         
$\;\;$ + SFT                   & 52.7              & 49.7            & \textbf{75.7}      & 37.2              & 42.2              & 44.3              & 50.3          \\ 
$\;\;$ + CP + SFT              & 54.3              & \textbf{50.6}   & 74.7               & 46.8              & 41.3              & \textbf{50.0}     & \textbf{53.0} \\
$\;\;$ + CP + SFT + DPO        & \textbf{54.4}     & 50.4            & 74.8               & 46.8              & 39.5              & 49.2              & 52.5          \\
$\;\;$ + CP + SFT + DPO + CoT  & 54.0              & 50.3            & 73.3               & \textbf{48.9}     & \textbf{43.7}     & 45.1              & 52.6          \\ \midrule

Mistral 7b                    & 39.3              & 36.8           & 76.3                & 24.5              & 31.2              & 27.9              & 39.3          \\
$\;\;$ + CP                   & 40.5              & 37.2           & 74.9                & 29.8              & 33.9              & 29.5              & 41.0          \\
$\;\;$ + SFT                  & 49.7              & 59.2           & 77.1                & \textbf{60.6}     & \textbf{66.1}     & 56.6              & \textbf{61.6} \\
$\;\;$ + CP + SFT             & \textbf{51.5}     & \textbf{60.9}  & 76.5                & 55.3              & 65.1              & 57.4              & 61.1          \\          
$\;\;$ + CP + SFT + DPO       & 49.3              & 57.3           & \textbf{77.3}       & 56.4              & 62.4              & 54.9              & 59.6          \\
$\;\;$ + CP + SFT + DPO + CoT & 51.0              & \textbf{60.9}  & 63.5                & 59.6              & 59.6              & \textbf{63.9}     & 59.8          \\

\bottomrule
\end{tabular}
}
\caption{\textbf{\HippoL~and \HippoM: Analysis of Continued Pretraining, Instruction Tuning, and Direct Preference Optimization.} This table demonstrates the incremental impact of Continued Pretraining (CP) on medical text data, Instruction Tuning (SFT), and Direct Preference Optimization (DPO) on the zero-shot capabilities of the LLaMA2 7B and Mistral 7B models across a range of medical benchmarks, including MedMCQA, MedQA, PubmedQA, and the USMLE series. The results, aggregated and individual, underline the significance of each methodological advancement in enhancing the model's proficiency in interpreting and responding to complex medical queries, thereby providing a granular view of performance improvements at each stage of model optimization.}
\label{tab: hippollama-hippomistral}
\end{table*}



\paragraph{\HippoL.} Our evaluation methodology for the LLaMA2 7B model covers successive training stages: Continued Pre-training (CP), Instruction Tuning (SFT), and Direct Preference Optimization (DPO). As listed in Table \ref{tab: hippollama-hippomistral}, the base model LLaMA2 7B initially achieves an average accuracy of 34.0 across benchmarks. The CP stage marginally increases accuracy to 34.4, indicating initial benefits from domain-focused continued pre-training. The subsequent introduction of SFT yields a substantial performance boost to an average accuracy of 50.3, demonstrating the critical role of customized instruction in enhancing the model’s capabilities in understanding and answering medical queries. Integrating CP with SFT further improves this performance to 53.0, highlighting the combined value of domain knowledge and specific instruction tuning. The final DPO stage slightly decreases the model's performance to 52.5, albeit with a slight increase in accuracy for MedMCQA and PubMedQA, illustrating DPO's refined impact on model preference alignment. This sequence delineates the incremental enhancements attributable to each training phase, with SFT marking a pivotal improvement. The composite model, LLaMA2 + CP + SFT, is thus designated as \HippoL~for its distinguished performance across our benchmarks.

\paragraph{\HippoM.} Following the approach for \HippoL, the training evolution for the Mistral 7B model reveals gradual improvement in the model's proficiency in medical question-answering. Initial results from the baseline Mistral 7B model, as shown in Table \ref{tab: hippollama-hippomistral}, show an average benchmark accuracy of 39.3. Implementing CP slightly improves this to 41.0, reflecting the positive yet modest impact of domain-specific continued pre-training. The pivotal SFT stage significantly raises the performance, achieving an average accuracy of 61.6, emphasizing the critical role of customized instruction in enhancing the model's interpretative and response capabilities for medical inquiries. Interestingly, combining CP and SFT results in a slight reduction to 61.1, suggesting a complex interaction between domain pre-training and instruction tuning. The subsequent application of DPO slightly lowers the overall score to 59.6, similar to the pattern observed for \HippoL, with targeted performance adjustment. Based on comprehensive analysis, Mistral 7b + SFT is selected to represent \HippoM, credited for its exceptional performance across all benchmarks.


\subsection{Chain-of-Thought (CoT) Prompting} The CoT prompting technique~\citep{wei2023chainofthought} enhances an LLM's ability to tackle complex queries by guiding it to articulate intermediate reasoning steps. This method improves the model's responses by structuring its problem-solving process. In our study, we applied CoT prompting for in-context learning, adopting a slightly altered instruction utilized in~\citep{pal2024gemini}: "\textit{The following is a multiple choice question about medical knowledge. Solve it in a step-by-step fashion, starting by summarizing the available information. Output a single option from the four options as the final answer.}". However, the application of CoT prompting in our experiments with downstream medical tasks did not consistently enhance our models' performance, as shown in Table~\ref{tab: hippollama-hippomistral}.


\subsection{Influencing Examples}
\label{sec: influence-examples}
We explore the application of Influence Functions to understand the behavior of LLMs~\citep{grosse2023studying} -- in our context, particularly those trained with domain-specific datasets like medical text. This technique quantifies the effect of single training instances on the model's predictions, improving the transparency of the AI models. This is increasingly important as the field of Explainable AI (XAI) grows to make AI systems more interpretable and accountable. However, the complexity of LLMs, which process vast amounts of data, highlights the necessity for efficient methods to perform this analysis. We believe incorporating this tool to our evaluation framework will prove useful for future studies. 

In the supplementary material (Appendix~\ref{sec:influence-appendix}), we present our analysis results, highlighting the most and least influential training examples for a MedQA dataset question and its model response. Notably, the most influential example shares overlapping medical concepts, in contrast to no shared concepts with the least influential training example.

\subsection{Uncertainty Quantification}
\label{sec: uncertainty-quantification}
\begin{figure}[!h]
\centering
\includegraphics[width=\linewidth]{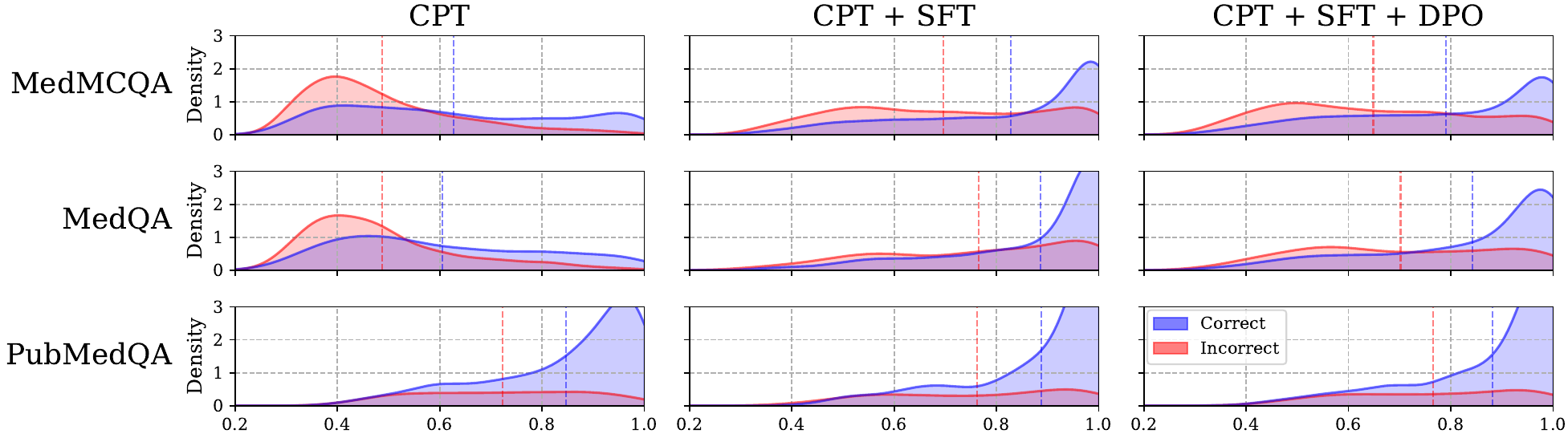}
\caption{\textbf{Uncertainty quantification for our best-performing 5-shot  \HippoM~model.}, where we plot the probability distributions assigned by the model to both correct predictions and incorrect predictions on the MedMCQA, MedQA, and PubMedQA datasets.}
\label{fig:uncertainty}
\end{figure}

In our study, we conducted an uncertainty quantification experiment on \HippoM~to understand its performance on the MedMCQA, MedQA, and PubMedQA datasets, as shown in Fig.\ref{fig:uncertainty}. Our findings reveal that our model consistently assigns higher probabilities to questions it answers correctly across all datasets, suggesting an ability to self-calibrate its certainty. The model’s confidence is notably higher on MedMCQA, possibly reflecting the dataset's relative simplicity. In contrast, its confidence on PubMedQA is comparatively lower, likely due to the dataset's complexity. Additionally, the model's confidence changes with different training stages; CPT leads to more conservative estimates, SFT boosts confidence, and adding DPO leads to variable confidence, with noticeable effects in MedMCQA and MedQA. These outcomes emphasize a complex relationship between training approaches and confidence calibration in the model.

We present additional negative results in Appendix~\ref{sec: negative-results}, which we anticipate will be beneficial for the community. By sharing these findings, we aim to encourage further investigations.

\section{Conclusion}
\label{sec: conclusion}
In this study, we have introduced Hippocrates, a comprehensive and open-source framework tailored for the medical domain, addressing a wide array of challenges faced by medical LLMs. We provide openly available datasets and establish an intuitive benchmark using the LM-Evaluation-Harness tool. We also introduce \HippoL~and \HippoM, two 7B models demonstrating superior performance. Our work makes substantial contributions to the field by combining in-depth empirical research with a structured training methodology, offering invaluable insights and tools for future research not only in healthcare but in any area requiring domain-specific adaptation of LLMs.

\bibliography{colm2024_conference}
\bibliographystyle{colm2024_conference}

\clearpage
\appendix
\section*{Appendix Table of Contents}
\noindent 
\begin{enumerate}
    \item \textbf{Appendix A}: Related Work
    \item \textbf{Appendix B}: Limitations and Safety
    \item \textbf{Appendix C}: Medical Large Language Models
    \item \textbf{Appendix D}: Evaluation Datasets
    \item \textbf{Appendix E}: Rationale for Selecting Mistral 7B and LLaMA2 7B as Base Models
    \item \textbf{Appendix F}: Data Contamination Analysis
    \item \textbf{Appendix G}: Additional Details on Training Stages
    \item \textbf{Appendix H}: Influence Functions
    \item \textbf{Appendix I}: Evaluation
    \item \textbf{Appendix J}: Negative Results 
\end{enumerate}

\section{Related Work}
\label{sec:related-work}

\noindent\textbf{Large Language Models.} The evolution of LLMs has marked a significant milestone in the field of NLP, with key developments including scaling efforts \citep{gpt32020, gpt4, palm22023, llama22023, gemini}. Meta-AI's introduction of the LLaMa base models \citep{llama2023, llama22023} and Ali Baba's 
 Qwen models \citep{Bai2023QwenTR} challenged the prevalence of closed models such as those from OpenAI \citep{gpt4} by adopting an open-source philosophy, thus democratizing access to state-of-the-art LLMs. This shift encouraged the community to engage in fine-tuning these base models with instructional datasets \citep{alpaca} and exploring self-instructional techniques, resulting in noticeable improvements across both quantitative and qualitative evaluations \citep{orca2023, platypus2023}. The application of parameter-efficient fine-tuning methods \citep{lora2021, qlora2023} addressed computational constraints, facilitating the development of domain-specific models, including those for medical applications \citep{chatdoctor2023, medalpaca2023, pmcllama2023, clinicalcamel2023, meditron70b}.

\noindent\textbf{Medical Large Language Models.} Models like ChatDoctor \citep{chatdoctor2023}, fine-tuned on 100,000 physician-patient dialogues using the LLaMA architecture \citep{llama2023}, have shown superior performance in medical QA-generation over GPT-3.5. Similarly, MedAlpaca \citep{medalpaca2023} leverages the LLaMA \citep{llama2023} model with LoRA \citep{lora2021} fine-tuning on 160,000 medical entries, demonstrating its efficacy on the USMLE self-assessment test. PMC-LLaMA \citep{pmcllama2023} applies LoRA fine-tuning on 4.8 million biomedical articles, while Clinical-Camel, utilizing the QLoRA \citep{qlora2023} tuning approach on LLaMA-2 \citep{llama22023}, sets new benchmarks in medical LLM performance. MEDITRON \citep{meditron70b} represents a significant advancement, training on a vast medical corpus with variations including 7B and 70B parameter models, indicating the potential of comprehensive datasets in enhancing LLM performance.  Further details regarding medical LLMs can be found in Appendix \ref{sec: llms}.


\noindent\textbf{Preference Learning.} Reinforcement learning from human feedback (RLHF) represents a methodology employed in the training of machine learning models to ensure their alignment with human objectives \citep{rlhf0, rlhf1, rlhf2}. From this line, RLHF has become the predominant approach for refining cutting-edge LMs \citep{instructgpt, gpt4, llama22023}. RLHF contains three main steps: collecting comparison data from human feedback, training a reward model (RM) on the comparison data, and learning a policy to maximize the reward with RL \citep{rlhf2, instructgpt}. Learning an RM from human feedback is not an easy procedure where you need to tune your LLM via RL by maximizing the predicted rewards under the supervision of your RM, while staying relatively close to the original model. Moreover, training with PPO can sometimes be unstable and requires careful hyper-parameter tuning. On the other hand, Direct Preference Optimization (DPO) \cite{dpo} offers a stable, efficient, and computationally light approach for fine-tuning LLMs to better align with human preferences, without the need for sampling from the LLM or extensive hyperparameter tuning. Moreover, as preference learning necessitates manually annotated human preferences, acquiring these annotations can be challenging or prohibitively expensive. Methods such as RLAIF \cite{rlaif} streamline this process by employing expert models that annotate using detailed prompts, thus reducing reliance on human annotators. This approach was also adopted in the creation of our doctor preference dataset, which was subsequently validated by actual medical doctors. 

\section{Limitations and Safety}

\paragraph{Model Limitations.} While our 7B model has achieved state-of-the-art results within its class, it is important to acknowledge its limitations compared to larger models such as OpenAI's GPT4 \citep{gpt4-medical-capabilities, medgemini}. The constraints imposed by the smaller parameter size may impede the model's reasoning capabilities, a crucial aspect of complex medical decision-making. Additionally, the model's performances are almost half on the average which highlights a huge area for improvement in open-source models.

\paragraph{Safety and Risks.} Crucially, despite these advancements, it is important to highlight that these AI models need substantial improvements before they can be safely and effectively employed with real patients. They are not yet at a stage where they can provide medical advice or be utilized for commercial healthcare applications. This limitation highlights the need for ongoing, careful development and validation of AI systems to guarantee their reliability and safety in clinical settings. The path toward AI integration in patient care is still unfolding, and while it holds promise, it requires a methodical and thoroughly evaluated approach.

\section{Medical Large Language Models}
\label{sec: llms}
\paragraph{BioMedLM.} BioMedLM\footnote{\url{https://crfm.stanford.edu/2022/12/15/biomedlm.html}}\footnote{\url{https://www.mosaicml.com/blog/introducing-pubmed-gpt}} is a decoder-based LLM with 2.7 billion parameters which was developed in the style of GPT \citep{gpt2} and trained on biomedical abstracts and papers. Model weights are available at Huggingface\footnote{\url{https://huggingface.co/stanford-crfm/BioMedLM}}.

\paragraph{BioGPT.} BioGPT \citep{biogpt2022} is a specialized LLM for generating and analyzing biomedical texts. BioGPT is built on  GPT-2 architecture \citep{gpt2} and has been trained from scratch using 15 million PubMed abstracts. BioGPT comes in two variants: the first is based on the GPT-2 medium model, while the second, BioGPT-Large, is built upon the GPT-2 XL, the largest version of GPT-2. Both versions are available at Huggingface\footnote{\url{https://huggingface.co/microsoft/biogpt}}\footnote{\url{https://huggingface.co/microsoft/BioGPT-Large}}. During our evaluations, we used BioGPT-Large.

\paragraph{ChatDoctor.} ChatDoctor \citep{chatdoctor2023} is a medical assistant LLM that is built on the LLaMA-7B \citep{llama2023} architecture and further refined using a comprehensive dataset of 100,000 patient-doctor interactions. Additionally, a separate dataset comprising 10,000 patient-doctor conversations from \url{iCliniq.com} has been released for testing. All resources including the model weights\footnote{\url{https://huggingface.co/zl111/ChatDoctor}} and datasets\footnote{\url{https://github.com/Kent0n-Li/ChatDoctor}} are publicly available.

\paragraph{ClinicalCamel.} Clinical Camel \citep{clinicalcamel2023} is a specialized open medical LLM based on the LLaMA-2-70B \citep{llama22023} architecture, enhanced with QLoRA \citep{qlora2023} for medical and clinical research applications. It is fine-tuned on three different data sources: ShareGPT conversations, 100,000 synthetic dialogues from Clinical Articles, and 10,187 USMLE questions from MedQA \citep{medqa}. Dialogues were generated using dialogue-based knowledge encoding (DBKE) in conjunction with questions from MedQA. Model weights \footnote{\url{https://huggingface.co/wanglab/ClinicalCamel-70B}} and the evaluation code \footnote{\url{https://github.com/bowang-lab/clinical-camel}} are publicly available.

\paragraph{MedAlpaca.} MedAlpaca \citep{medalpaca2023} introduces a unique IT dataset with over 160,000 entries, designed for optimizing LLMs for medical uses. The researchers focused on SFT training for the 7B and 13B variants of LLaMA \citep{llama2023, llama22023}. They also developed an evaluation approach based on the models' zero-shot performance on self-assessment datasets from Steps 1, 2, and 3 of the United States Medical Licensing Examination (USMLE). All the code\footnote{\url{https://github.com/kbressem/medAlpaca}}, datasets\footnote{\url{https://github.com/kbressem/medAlpaca/blob/main/DATA_DESCRIPTION.md}}, and model variants\footnote{\url{https://huggingface.co/medalpaca}} are publicly available.

\paragraph{PMC-LLaMA.} PMC-LLaMA \citep{pmcllama2023} is built upon on an integration of a vast amount of medical data pre-training and IT datasets. Its pre-training data includes 4.8 million biomedical academic papers and 30,000 medical textbooks. On the other hand, they have developed a large-scale dataset for instruction tuning which encompasses various components like medical question-answering, rationales for reasoning, and conversational dialogues, totaling 202 million tokens. They introduced two distinct models: MedLLaMA, trained exclusively on the pre-training dataset, and PMC-LLaMA, which underwent training with both the pre-training and instruction tuning (IT) datasets based on the LLaMA2-7B and LLaMA2-13B architectures \citep{llama22023}. The code\footnote{\url{https://github.com/chaoyi-wu/PMC-LLaMA}} and the model weights\footnote{\url{https://huggingface.co/axiong/PMC_LLaMA_13B}} are publicly available.

\paragraph{Meditron.} Meditron \citep{meditron70b} is a recent study that adapts two large-scale medical LLMs from Llama-2, Meditron-7B and Meditron-70B. These models have undergone additional training via Megatron-LM \citep{megatronllm} on a specially curated medical corpus. This corpus includes selected PubMed papers and abstracts, a new dataset of internationally recognized medical guidelines, and a general domain corpus. The code\footnote{\url{https://github.com/epfLLM/meditron}}, dataset\footnote{\url{https://huggingface.co/datasets/epfl-llm/guidelines}}, and model weights\footnote{\url{https://huggingface.co/epfl-llm}} are publicly available.

\paragraph{BioMistral.} BioMistral \citep{biomistral} is a recent open-source LLM developed specifically for the biomedical domain, built upon the Mistral foundation model and enriched through pre-training on PubMed Central. This model distinguishes itself through comprehensive evaluation across 10 established medical question-answering tasks in English, with additional exploration into multilingual applications by translating these tasks into 7 other languages. The creation of BioMistral, including its derivative models through quantization and novel model merging techniques, exemplifies a leap in blending specialized and general-purpose LLM capabilities, notably in terms of medical accuracy and multilingual robustness.

\section{Evaluation Datasets} 
MedMCQA \citep{medmcqa}, derived from Indian medical entrance exams such as AIIMS and NEET-PG, includes over 194,000 high-quality multiple-choice questions spanning 2,400 healthcare topics across 21 medical subjects. This benchmark includes 4183 test samples\footnote{\url{https://huggingface.co/datasets/medmcqa}}. 
MedQA \citep{medqa}, created from the United States Medical License Exams (USMLE), aggregates a broad spectrum of professional board examination questions, presented in a multiple-choice format with four options. For our analysis, following prior works, we used several test splits of MedQA provided by Huggingface\footnote{\url{https://huggingface.co/augtoma}}. 
PubMedQA \citep{pubmedqa}, a biomedical question-answering dataset, is sourced from PubMed abstracts and aims to answer research questions as yes, no, and maybe options. It consists of three subsets PQA-artificial, PQA-labeled, and PQA-unlabeled.

\section{Rationale for Selecting Mistral 7B and LLaMA2 7B as Base Models}
\label{sec: llm-selection}
In developing a comprehensive framework for medical LLMs, we explored the integration of two prominent models, Mistral 7B and LLaMA2 7B, to enhance the robustness and versatility of our system. This decision was motivated by the distinct architectural nuances and training paradigms inherent to each model, and the promising baseline results of these models in our evaluations. Our framework's dual-model approach aims to capitalize on the strengths of both LLaMA2 7B and Mistral 7B, acknowledging that the efficacy of fine-tuning strategies can vary significantly across different base models. By incorporating Meditron, which builds upon LLaMA's architecture and training insights, we ensure a robust baseline for medical language tasks. Concurrently, the inclusion of Mistral allows for comparative analysis, enriching our understanding of model behavior in the face of medical data intricacies.

\section{Data Contamination Analysis} 
\label{sec: data-contamination-analysis}

Data contamination poses a significant concern for LLMs, as their extensive training corpora can obscure the presence of data leakages. To ensure the integrity and unbiased assessment of LLMs, the evaluation benchmarks must remain uncontaminated by the training datasets. In our study, we examined the n-gram overlap between the test splits of evaluation benchmarks and the three distinct categories of training sets: Continued Pre-training Data, General Instruction Data, and Evaluation Instruction Data. Our method involved calculating the overlap of 3-grams and 5-grams between the evaluation benchmarks and these training datasets, adopting OpenAI's preprocessing technique, which includes the removal of punctuation, symbols, and normalization of case sensitivity \citep{gpt4}. While previous studies analyzed higher n-gram numbers, such as 8-grams in Meditron \citep{meditron70b} and 13-grams in GPT-3 \cite{gpt3}, our investigation did not extend beyond 5-grams. This decision was based on the observation that the overlap ratio at the 5-gram level was minimal and thus negligible. The outcomes for each evaluation benchmark are detailed separately in Table \ref{tab: sft-overlap}.
\begin{table}[h]
\resizebox{\linewidth}{!}{
\begin{tabular}{@{}l@{$\quad$}c@{$\quad$}c@{$\quad$}c@{$\quad$}c@{$\quad$}c@{$\quad$}c@{}}
\toprule
\multirow{2}{*}{\textbf{Dataset}} & \textbf{MedMCQA} & \textbf{MedQA} & \textbf{PubmedQA} & \textbf{USMLE-1} & \textbf{USMLE-2} & \textbf{USMLE-3} \\ 
 & $n=3$/$n=5$ & $n=3$/$n=5$ & $n=3$/$n=5$ &  $n=3/n=5$ & $n=3$/$n=5$ & $n=3/n=5$ \\ 
\toprule
Medical Guidelines & 0.10 / 0.01 & 0.16 / 0.01 & 0.26 / 0.02 & 0.01 / 0.00 & 0.02 / 0.00 & 0.02 / 0.00 \\
PMC-Patients & 0.10 / 0.01 & 0.23 / 0.03 & 0.22 / 0.01 & 0.02 / 0.00 & 0.03 / 0.00 & 0.03 / 0.00 \\
PubMedQA-train & 0.09 / 0.00 & 0.12 / 0.01 & 0.37 / 0.05 & 0.01 / 0.00 & 0.02 / 0.00 & 0.02 / 0.00 \\
\midrule
Medical Flashcards & 1.08 / 0.05 & 1.65 / 0.10 & 1.55 / \textbf{0.07} & 0.16 / 0.01 & 0.20 / 0.01 & 0.25 / 0.01 \\ 
GenMedGPT-5k & 1.43 / 0.03 & 3.02 / 0.13 & 1.67 / 0.03 & 0.34 / 0.01 & 0.39 / 0.01 & 0.55 / 0.01 \\ 
Open Platypus & 0.27 / 0.01 & 0.36 / 0.01 & 0.50 / 0.02 & 0.04 / 0.00 & 0.05 / 0.00 & 0.07 / 0.00 \\ 
HealthCareMagic-100k & 0.20 / 0.01 & 0.40 / 0.03 & 0.28 / 0.01 & 0.03 / 0.00 & 0.05 / 0.00 & 0.06 / 0.00 \\ 
UMLS & 0.92 / 0.03 & 1.13 / 0.06 & 1.41 / 0.06 & 0.13 / 0.01 & 0.15 / 0.00 & 0.19 / 0.01 \\ 
UMLS-Relation & 0.52 / 0.00 & 0.66 / 0.00 & 0.61 / 0.00 & 0.07 / 0.00 & 0.11 / 0.00 & 0.12 / 0.00 \\ 
Wikidoc & 0.95 / 0.00 & 1.37 / 0.00 & 1.83 / 0.00 & 0.14 / 0.00 & 0.18 / 0.00 & 0.24 / 0.00 \\ 
Wikidoc-Patient-Info & 1.45 / 0.00 & 2.33 / 0.00 & 2.29 / 0.00 & 0.25 / 0.00 & 0.34 / 0.00 & 0.44 / 0.00 \\ 
MedicationQA & \textbf{3.11} / 0.00 & 4.08 / 0.00 & \textbf{5.24} / 0.00 & 0.52 / 0.00 & 0.65 / 0.00 & 0.97 / 0.00 \\ 
\midrule
MedMCQA-train & 1.85 / \textbf{0.37} & 1.52 / 0.24 & 0.69 / 0.02 & 0.15 / 0.03 & 0.19 / 0.03 & 0.23 / 0.03 \\
MedQA-train & 1.71 / 0.12 & \textbf{7.61} / \textbf{2.78} & 1.51 / 0.05 & \textbf{0.60} / \textbf{0.18} & \textbf{0.90} / \textbf{0.25} & \textbf{0.99} / \textbf{0.22} \\
PubMedQA-train & 0.08 / 0.00 & 0.12 / 0.01 & 0.35 / 0.04 & 0.01 / 0.00 & 0.01 / 0.00 & 0.02 / 0.00 \\
\bottomrule
\end{tabular}%
}
\caption{Comparative Analysis of N-gram Overlap Ratios between Evaluation Benchmark Datasets and Various Training Sets: Continued Pre-training Data, General Instruction Data, and Evaluation Instruction Data. The overlap ratio is determined by dividing the count of matching n-grams found in both the evaluation benchmark and a training dataset by the total count of n-grams in the evaluation benchmark.}
\label{tab: sft-overlap}
\end{table}

\section{Additional Details on Training Stages}
\label{sec: appendix-training-stages}
\paragraph{Continued Pre-training (CP)}
\label{sec: appendix-continued-pre-training}
Each medical text corpus originates from diverse sources, with each potentially exhibiting different distributions despite containing medical information. To evaluate the effectiveness of each corpus, we conducted various experiments involving both individually and by combining them. During these experiments, we utilized Low-Rank Adaptation (LoRA) with $r=8$ and $\alpha=16$, targeting the fully-connected layers of the corresponding LLM \cite{lora2021}. We set the learning rate at $1e-4$ for each experiment, employing a batch size of 8 with two gradient accumulations, and go through all tokens in the input dataset exactly once. To enhance training convergence, we applied a cosine scheduler alongside the Adam optimizer. Our comprehensive experiments, detailed in Table \ref{tab: llama-experiments} and utilizing the base LLM LLaMA2 7B \cite{llama22023}, show that each data split contributes differently to downstream task accuracy. Notably, the highest scores were achieved using a combination of the PubMedQA-train split \cite{pubmedqa}, which provides context for each sample, and Medical Guidelines from the Meditron dataset \cite{meditron70b}. These findings remained consistent when employing the Mistral 7B as the base model \cite{mistral7b}.

\paragraph{Supervised Finetuning (SFT)}
\label{sec: appendix-sft}
The instruction Tuning (IT) stage is one of the tricky steps to enhance the knowledge and reasoning capabilities of LLMs one step forward. To do that, we gathered lots of IT  datasets from different sources. We categorize our IT datasets as two distinct groups: (i) General Instruction Datasets and (ii) Evaluation Instruction Datasets. General Instructions contains over 400,000 samples from nine different datasets, originating from the instruction corpora of previous studies like MedAlpaca \cite{medalpaca2023}, PMC-LLaMA \cite{pmcllama2023}, and Platypus \cite{platypus2023}. It aims to reduce bias and enhance generalization across various reasoning tasks by excluding data from training or test splits of downstream QA benchmarks. A pre-processing protocol was used to eliminate irrelevant content, like unnecessary words and web URLs, to ensure the quality and relevance of the data. Detailed statistics of the dataset can be found in Table \ref{tab: dataset-instruction}. On the other hand, Evaluation Instruction Datasets are generated from the training splits of the downstream tasks, if training splits exist. It aimed to study the impact of incorporating instruction samples from downstream tasks to see the effect it by following similar works like Meditron \cite{meditron70b}, see Table \ref{tab: dataset-evaluation} for details. Instruction-response pairs were generated using the training splits of different benchmarks, based on the instruction templates provided by Meditron. A variety of experiments were conducted to assess the impact of each split on tasks, individually and collectively. Following an in-depth analysis of each split for both LLaMA2 7B and Mistral 7B, it was found that the concatenation of each Evaluation Instruction split yielded the highest scores in LLaMA. Conversely, for Mistral, the MedQA's Instruction split delivered the most superior outcomes. For additional insights into how each dataset influences downstream results, please see Table \ref{tab: llama-experiments}.


\section{Influence Functions}
\label{sec:influence-appendix}
To assess the influence of specific training examples on model parameters, we measure how changes in these parameters affect the model’s output when a data point is added, removed, or weighted differently in the training set, requiring the computation of the inverse-Hessian-vector product (IHVP) \citep{koh2020understanding}. Our approach to influence analysis combines the preliminary subsampling of training examples based on TF-IDF vector similarity with the application of DataInf, a computationally efficient method for calculating influence scores~\citep{kwon2024datainf}. We experiment with the MedQA dataset where we calculate the influence of training samples over test samples. Due to computational constraints mentioned in Appendix \ref{appendix:negative_res_analysis}, we subsample the MedQA train split to 1000 examples by selecting the most similar training samples compared to test samples based on a TF-IDF similarity, similar to \cite{grosse2023studying}.

In Figure~\ref{fig:influence_table}, we present an illustrative example showing the most and least influential MedQA instruction-tuning samples identified by our approach for a specific MedQA test example. The most influential training example is found to be similar to the test example, as measured in terms of the overlapping medical terms extracted by the MetaMap tool\footnote{https://metamap.nlm.nih.gov/}, contrasting sharply with the absence of any such overlap for the least influential example.

\begin{table*}[h]
\resizebox{\linewidth}{!}{
\begin{tabular}{lc@{$\;\;$}c@{$\;\;$}c@{$\;\;$}c@{$\;\;$}c@{$\;\;$}c@{$\quad$}c}
\toprule
\textbf{Experiment}                            & \textbf{MedMCQA}  & \textbf{MedQA}  & \textbf{PubmedQA}  & \textbf{USMLE-1}  & \textbf{USMLE-2}  & \textbf{USMLE-3}  &\textbf{Avg.}  \\ \midrule           

LLaMA2 7b                                 & 34.4              & 29.3            & 72.3               & 18.1              & 22.9              & 27.1              & 34.0          \\  \midrule

$\;\;$ + Guidelines                            & 35.2              & 31.7            & 70.6               & 20.2              & 22.9              & 24.6              & 34.2          \\
$\;\;$ + PMC                           & 34.2              & 30.9            & 71.1               & 9.6               & 20.2              & 24.6              & 31.8          \\
$\;\;$ + PubMedQA                           & 34.3              & 29.9            & 73.3               & 14.9              & 21.1              & 27.9              & 33.6          \\
$\;\;$ + PubMedQA  + Guidelines (PMQA + GDL)                     & 34.6              & 31.9            & 72.8               & 20.2              & 25.7              & 21.3              & 34.4          \\ 
$\;\;$ + PMC + Guidelines                      & 35.2              & 31.7            & 69.5               & 14.9              & 23.9              & 26.2              & 33.6          \\ 
$\;\;$ + PMC + Guidelines + PubMedQA                & 34.8              & 31.5            & 72.2               & 17.0              & 23.9              & 24.6              & 34.0          \\ \midrule           

$\;\;$ + General Instructions                         & 35.8              & 35.0            & 73.1               & 26.6              & 24.8              & 32.8              & 38.0          \\
$\;\;$ + PubMedQA Instruction                          & 32.2              & 27.5            & 55.2               & 22.3              & 17.4              & 18.0              & 28.8          \\
$\;\;$ + MedMCQA Instruction                        & 53.3              & 47.2            & 74.5               & 40.4              & 37.6              & 45.9              & 49.8          \\
$\;\;$ + MedqQA Instruction                      & 38.9              & 46.6            & 75.5               & 37.2              & 36.7              & 43.4              & 46.4          \\
$\;\;$ + Evaluation Instructions                          & 52.7              & 49.7            & 75.7               & 37.2              & 42.2              & 44.3              & 50.3          \\
$\;\;$ + General + Evaluation Instructions                 & 48.0              & 43.9            & 75.3               & 30.9              & 33.0              & 41.0              & 45.4          \\ \midrule

$\;\;$ + PMQA + GDL + General Instructions             & 35.2              & 32.5            & 72.2               & 20.2              & 27.5              & 23.8              & 35.2          \\
$\;\;$ + PMQA + GDL + PubMedQA Instructions             & 26.5              & 27.8            & 78.0               & 18.1              & 23.9              & 27.1              & 33.6          \\
$\;\;$ + PMQA + GDL + MedMCQA Instructions          & 54.6              & 45.8            & 74.6               & 36.2              & 43.1              & 47.5              & 50.3          \\
$\;\;$ + PMQA + GDL + MedqQA Instructions            & 39.0              & 47.5            & 73.5               & 41.5              & 33.0              & 42.6              & 46.2          \\
$\;\;$ + PMQA + GDL + Evaluation Instructions             & 54.3              & 50.6            & 74.7               & 46.8              & 41.3              & 50.0              & 53.0          \\ 
$\;\;$ + PMQA + GDL + General + Evaluation Instr.    & 47.6              & 41.2            & 75.0               & 38.3              & 34.9              & 37.7              & 45.8          \\ \midrule

$\;\;$ + PMQA + GDL + Evaluation Instr.  + DPO         & 54.4              & 50.4            & 74.8               & 46.8              & 39.5              & 49.2              & 52.5          \\ 

\midrule

Mistral 7B                                & 39.3              & 36.8            & 76.3                & 24.5             & 31.2              & 27.9              & 39.3          \\  \midrule
$\;\;$ + Guidelines                            & 41.4              & 38.4            & 74.8               & 28.7              & 31.2              & 31.2              & 40.9          \\
$\;\;$ + PMC                           & 40.0              & 37.9            & 75.2               & 25.5              & 30.3              & 33.6              & 40.4          \\
$\;\;$ + PubMedQA                           & 39.0              & 36.2            & 78.2               & 26.6              & 27.5              & 32.0              & 39.9          \\
$\;\;$ + PubMedQA  + Guidelines (PMQA + GDL)                     & 41.0              & 37.9            & 76.8               & 24.5              & 34.9              & 31.2              & 41.0          \\ 
$\;\;$ + PMC + Guidelines                       & 40.2              & 37.2            & 74.6               & 27.7              & 33.0              & 31.2              & 40.6          \\

$\;\;$ + PMC + Guidelines + PubMedQA                & 40.0              & 36.5            & 76.3               & 31.9              & 30.3              & 28.7              & 40.6          \\ \midrule
  
$\;\;$ + PubMedQA Instructions           & 32.5             & 29.5           & 60.4              & 19.2             & 18.4             & 31.2             & 31.9         \\
$\;\;$ + MedQA Instructions             & 49.7             & 59.2           & 77.1              & 60.6             & 66.1             & 56.6             & 61.6         \\ 
$\;\;$ + MedMCQA Instructions             & 60.6             & 53.1           & 75.5              & 58.5             & 47.7             & 52.5             & 58.0         \\ 
$\;\;$ + General Instructions         & 59.1             & 54.6           & 60.8              & 54.3             & 52.3             & 51.6             & 55.5         \\
$\;\;$ + General + Evaluation Instructions          & 51.2             & 48.9           & 75.6              & 52.1             & 37.6             & 47.5             & 52.1         \\  \midrule

$\;\;$ + PMQA + GDL + General Instructions       & 58.1             & 54.8           & 73.6              & 53.2             & 56.0             & 50.0             & 57.6         \\
$\;\;$ + PMQA + GDL + MedMCQA Instructions       & 57.4             & 50.0           & 78.1              & 45.7             & 45.0             & 52.5             & 54.8         \\
$\;\;$ + PMQA + GDL + MedQA Instructions       & 51.5             & 60.9           & 76.5              & 55.3             & 65.1             & 57.4             & 61.1         \\ 
$\;\;$ + PMQA + GDL + PubMedQA Instructions        & 32.7             & 29.2           & 68.2              & 14.9             & 29.4             & 34.4             & 34.8         \\
$\;\;$ + PMQA + GDL + General + Evaluation Instr.     & 50.7             & 48.2           & 76.6              & 48.9             & 38.5             & 44.3             & 51.2         \\ \midrule
         
$\;\;$ + PMQA + GDL + MedQA Instr. + DPO   & 49.3             & 57.3          & 77.3              & 56.4             & 62.4             & 54.9             & 59.6          \\  \bottomrule

\end{tabular}
}
\caption{LLaMA2 7B and Mistral 7B zero shot experiments on MedMCQA, PubMEDQA, MedQA, USMLE-step1, USMLE-step2, USMLE-step3 by using the LM-Evaluation-Harness.}
\label{tab: llama-experiments}
\end{table*}

\section{Evaluation}
\label{sec:appendix-eval}

\paragraph{Reliability.} To ensure a fair and easily replicable assessment of these medical models, we utilized the Eleuther AI Language Model Evaluation Harness \citep{eval-harness}, a unified evaluation framework specifically designed for evaluating generative LLMs. This framework is also the foundational evaluation tool for the Open LLM Leaderboard\footnote{\url{https://huggingface.co/spaces/HuggingFaceH4/open_llm_leaderboard}} \citep{open-llm-leaderboard}.

\paragraph{QA Evaluation Metric.} The LM-Evaluation-Harness operates on a Log-Likelihood objective, calculating the negative log-likelihood for each potential answer in response to a given query. The answer is then chosen based on the highest likelihood score, indicating it as the most probable choice. 

\paragraph{Prompting.} Our evaluation was conducted using widely recognized datasets from prior works. Specifically, we employed six different question-answering datasets: MedMCQA, MedQA, PubMedQA, USMLE-Step1, USMLE-Step2, and USMLE-Step3 \citep{medmcqa, medqa, pubmedqa}. Each prompt includes a question and corresponding choices, separated by a newline. For PubMedQA evaluations, we also incorporated the abstract as context for the model's reasoning. Prompt examples can be seen in Fig.~\ref{tab:appendix-medmcqa-medqa-pubmedqa} and Fig.~\ref{tab:appendix-usmle}.

\paragraph{Qualitative Results}
In Tables~\ref{tab:medmcqa_ai_responses} through \ref{tab:usmle_3_ai_responses}, we show representative samples from each benchmark employed in our evaluation. These tables include responses from our \HippoL~and \HippoM~models as well as from competing models. We use the LM-Evaluation-Harness for prompt formatting, and the results are obtained with zero-shot setting.

\section{Negative Results} 
\label{sec: negative-results}

This section points out the negative results associated with each section from the main text. The organization of these paragraphs closely reflects the structure of the main body.

\subsection{Hippocrates Framework}
\label{sec: negative-results-hippocrates-framework}

\paragraph{Combining Different Datasets.} In both Continued Pre-training and Supervised Fine-tuning, simply combining each dataset and adding more samples does not positively impact the downstream task. For Continued Pre-training, the best results were achieved by utilizing the Medical Guidelines and PubmedQA-train splits, while excluding the PMC-Patients dataset. On the other hand, during the process of instruction tuning, significant time was invested in compiling various IT datasets, as detailed in Table \ref{tab: dataset-instruction}. Furthermore, we created Evaluation Instruction datasets by adopting the prompts from Meditron \cite{meditron70b}. Despite possessing a vast array of instruction samples, including 292K from General Instructions, 182,822 MedMCQA-train samples, 10,178 MedQA-train samples, and 211,269 PubmedQA-train samples, our best performance was notably achieved by exclusively utilizing the MedQA-train split. Intriguingly, this split contained the smallest quantity of instruction samples yet yielded the most significant improvement by a considerable margin.

\paragraph{Impact of Continued Pre-training.} Continued Pre-training (CP) serves as a foundational step in customizing LLMs for domain-specific tasks. However, an examination of Table~\ref{tab: hippollama-hippomistral} shows that while Supervised Finetuning (SFT) improves the accuracy of LLaMA2 7B following CP; Mistral 7B, employing only SFT, outperforms the combination of CP followed by SFT. This suggests that the impact of CP prior to SFT may vary depending on the underlying base LLM.

\paragraph{Preference Dataset Creation.} We attempted to employ preference learning methods for the first time by utilizing RLAIF \cite{rlaif} for the medical domain. We created our comparison dataset using GPT-4 \cite{gpt4, gpt4-medical-capabilities}, which acted as the annotator. This approach was facilitated by providing a detailed instruction prompt (Figure \ref{fig:RLHAIF-prompt}), adapted from MedPaLM's instruction used by human annotators \cite{medpalm2022, medpalm22023}. Instead of creating a custom dataset comprising single prompts and their corresponding model outputs for GPT-4 to annotate based on specific instructions, we opted to directly leverage the iCliniq dataset. This dataset encompasses three distinct responses from a real doctor and other LLMs. This approach raises an open question: Could there be an improved alignment with medical preferences if our own LLMs generated the responses?

\paragraph{RLAIF with DPO.} We utilized the Medical Comparison Dataset outlined in Section \ref{sec: medical-comparison-data} to enhance the medical alignment by applying DPO \cite{dpo} to learn from medical feedback and update our models. However, as indicated in Table \ref{tab: hippollama-hippomistral}, training with DPO resulted in modest improvements for PubMedQA and certain USMLE steps for both LLaMA2 and Mistral-based models, but a slight decrease in overall performance across all tasks in the benchmark. We hypothesize that this may be due to the similarity in question format between our preference datasets and PubMedQA. Our datasets include a detailed explanation of the patient's current issue followed by a related question, mirroring PubMedQA's structure of presenting an abstract related to the problem before posing the question. In contrast, other datasets follow a more straightforward question-and-answer format.

\subsection{Analysis}
\label{appendix:negative_res_analysis}

\paragraph{Model Selection.} As previously noted in the Appendix \ref{sec: negative-results-hippocrates-framework} on negative results within the Hippocrates framework, the sequential application of CP, followed by SFT, and then Preference Learning, does not yield optimal models for LLaMA2 7B and Mistral 7B. Our top-performing model for LLaMA2 7B, dubbed \HippoL, was developed by CP, immediately followed by SFT. Conversely, for the Mistral 7B model, our best results, leading to the creation of \HippoM, were achieved solely through SFT.

\paragraph{Additional Prompting Strategies.} Our incorporation of
Chain-of-Thought (CoT) as an additional prompting strategy
resulted in slight improvements in certain subtasks but an overall decrease in accuracy (see Table \ref{tab: hippollama-hippomistral}), mirroring findings from PubMedQA CoT evaluations in MedPaLM \cite{medpalm2022, medpalm22023}. This underscores an intriguing avenue for further exploration, given the generally high accuracy these strategies motivate in LLMs \cite{wei2023chainofthought, wang2023selfconsistency}.

\paragraph{Influence functions.} The main challenge regarding influence functions was CPU memory limitations. In the case of \HippoL~, each example has approximately 200 MBs of gradients, therefore storing gradients for a total of 700000 examples needs 140 TB hard disk and RAM space, excluding any essential memory requirements. Therefore, we subsample the MedQA train split to 1000 examples and perform a small-scale qualitative analysis. 

\clearpage
\begin{figure}[!h]
\begin{tcolorbox}[colback=gray!5!white,colframe=black!95!black,title=\textbf{Test Sample}] 
\small \textbf{Question:} A 4-week-old female newborn is brought to the physician because of increasing yellowing of her eyes and skin for 2 weeks. The mother has noticed that the girl's stools have become pale over the past week. She was breastfed since birth but her parents switched her to formula feeds recently after reading on the internet that breastfeeding could be the cause of her current symptoms. The patient was delivered vaginally at 38 weeks' gestation. Pregnancy and delivery were uncomplicated. She appears healthy. Vital signs are within normal limits. She is at the 50th percentile for length and at the 60th percentile for weight. Examination shows scleral icterus and jaundice. The liver is palpated 2 cm below the right costal margin. Cardiopulmonary examination shows no abnormalities. Neurologic examination shows no focal findings. Serum studies show:
Bilirubin
Total 15 mg/dL
Direct 12.3 mg/dL
Alkaline phosphatase 2007 U/L
AST 53 U/L
ALT 45 U/L
$\gamma$-glutamyl transferase 154 U/L
Blood group A positive
Which of the following is the most likely diagnosis?" 
\\ \textbf{Options:} (A) Galactosemia
(B) Biliary atresia
(C) Crigler–Najjar syndrome
(D) Breast milk jaundice \textbf{Answer:} Biliary atresia 
\end{tcolorbox}
\begin{tcolorbox}[colback=gray!5!white,colframe=black!95!black,title=\textbf{Most Influential Training Sample}] 
\small \textbf{Question:} A 10-month-old girl is brought to the physician by her mother because of fever and irritability for the past 2 days. The mother says that the girl's diapers have smelled bad since the symptoms started. The patient has had some clear nasal secretions over the past week. Two months ago, she was brought to the emergency department for a simple febrile seizure. Otherwise, she has been healthy and her immunizations are up-to-date. She appears ill. She is at the 50th percentile for height and weight. Her temperature is 39.1°C (102.3°F), pulse is 138/min, respirations are 26/min, and blood pressure is 75/45 mm Hg. Oropharyngeal examination shows a mild postnasal drip. The remainder of the examination shows no abnormalities. Laboratory studies show:
Hemoglobin 12.4 g/dL
Leukocyte count
8,000/mm3
Serum
Na+ 138 mEq/L
K+ 4.0 mEq/L
Cl- 100 mEq/L
Creatinine 0.5 mg/dL
Urine
RBC 1–2/hpf
WBC 18–20 WBCs/hpf
Nitrites positive
Bacteria gram-negative rods
Nasal swab for respiratory syncytial virus, influenza A, and influenza B antigens is negative. Urine culture grows $>$ 105 colony forming units (CFU)/mL of E. coli. Treatment with acetaminophen and cefixime is started. Two days later, her symptoms have improved. Which of the following is the most appropriate next step in management?"
\\ \textbf{Options:}
(A) Obtain CT scan of the abdomen
(B) Perform renal and bladder ultrasound
(C) Perform an intravenous pyelogram (IVP)
(D) Start prophylaxis with trimethoprim-sulfamethoxazole
\textbf{Answer:} 
Perform renal and bladder ultrasound 
\end{tcolorbox}
\begin{tcolorbox}[colback=gray!5!white,colframe=black!95!black,title=\textbf{Least Influential Training Sample}] 
\small \textbf{Question:} A 35-year-old woman presents to the emergency room with chest pain. She describes the chest pain as severe, 9/10, sharp in character, and diffusely localized to anterior chest wall. She also says she is sweating profusely and feels like “she is about to die”. She has presented to at least 4 different emergency rooms over the past month with similar episodes which resolve after 10–15 minutes with no sequelae or evidence of cardiac pathology. However, she says she is fearful every day of another episode. No significant past medical history. Vital signs are within normal limits, and physical examination is unremarkable. Laboratory findings, including cardiac troponins, are normal. Which of the following is the best pharmacological treatment for long-term management of this patient?
\\ \textbf{Options:}
(A) Paroxetine
(B) Benzodiazepine
(C) Phenelzine
(D) Nortriptyline
\textbf{Answer: }
Paroxetine 
\end{tcolorbox}
\vspace{-0.3cm}
\caption{The most and least influential MedQA instruction-tuning samples for a MedQA test sample for the \HippoL~model. The test sample and the most influential sample are more similar compared to the least influential sample.}
\label{fig:influence_table}
\end{figure}

\begin{figure}[!h]
\begin{tcolorbox}[colback=gray!5!white,colframe=black!95!black,title=\textbf{\small{MedQA} | \textbf{Format:} Question}] 
\small \textbf{Question:} A 3-week-old newborn is brought to the physician by his parents because of poor feeding, irritability, and frequent vomiting over the past week. The vomitus is greenish in color and smells strange. His parents have tried to feed him every 4 hours, but the patient often spits up or refuses to eat. The patient was born at term and had his first bowel movement at 50 hours of life. He has since had one bowel movement daily. He is at the 50th percentile for length, 10th percentile for weight, and 40th percentile for head circumference. He does not appear to be in acute distress. His temperature is 36.9°C (98.4°F), pulse is 140/min, respirations are 40/min, and blood pressure is 90/60 mm Hg. Physical examination shows that the patient has small, low-set ears, a broad and flat nasal bridge, and a large space between the first and second toes bilaterally. The abdomen is distended. When the finger is removed following a rectal exam, there is an explosive release of stool from the patient's rectum. An x-ray of the abdomen shows a section of dilated colon followed by a segment of colon without stool or air.\\
    
    Which of the following is most likely to confirm the diagnosis?\\
    (A) CT scan of the abdomen (B) Transabdominal ultrasonography (C) Anorectal manometry (D) Rectal suction biopsy
\end{tcolorbox}

\begin{tcolorbox}[colback=gray!5!white,colframe=black!95!black,title=\textbf{\small{MedMCQA} | \textbf{Format:} Question}] 
\small \textbf{Question:} Tensor veli palatini is supplied by:\\

(A) Facial nerve (B) Trigeminal nerve (C) Glossopharyngeal nerve (D) Pharyngeal plexus

\end{tcolorbox}

\begin{tcolorbox}[colback=gray!5!white,colframe=black!95!black,title=\textbf{\small {PubMedQA} | \textbf{Format:} Abstract + Question}] 
\small \textbf{Abstract:} The use of open access endoscopy is increasing. Its effect on the adequacy of patient informed consent, procedure acceptance and the impact on subsequent communication/transfer of procedure results to the patient have not been evaluated. The aim of our study was to compare the extent of preknowledge of procedures and test explanation, patient medical complexity, information transfer and overall patient satisfaction between a patient group referred for outpatient open access endoscopy versus a patient group from a gastrointestinal (GI) subspecialty clinic. Information was obtained from all patients presenting for outpatient upper and lower endoscopy by using a 1-page questionnaire. Patients from the two groups who had an outpatient upper/lower endoscopic procedure were contacted by phone after the procedure to obtain information with a standardized questionnaire. The open access patients reported receiving significantly less information to help them identify the procedure (p$<$0.01) and less explanation concerning the nature of the procedure than the group of patients referred from the subspecialty clinic (p$<$0.005). There was no difference between the two groups in satisfaction scores for examinations performed under conscious sedation. For flexible sigmoidoscopy without sedation, however, the GI clinic patient group were more satisfied with their procedure. The majority of patients, regardless of access, were more likely to receive endoscopic results from a gastroenterologist than the referring physician. Furthermore, the patients in the GI clinic group who underwent colonoscopy felt significantly better at follow-up.\\

\textbf{Question:} Does open access endoscopy close the door to an adequately informed patient?                                                 (A) yes (B) no (C) maybe

\end{tcolorbox}
\vspace{-0.25cm}
\caption{Examples of prompts used in the evaluation of MedMCQA, MedQA, and PubMedQA. {\textbf{Format}} shows the information order in the prompt.}
\label{tab:appendix-medmcqa-medqa-pubmedqa}
\end{figure}
\begin{figure}[!h]
\begin{tcolorbox}[colback=gray!5!white,colframe=black!95!black,title=\textbf{\small{USMLE-Step1} | \textbf{Format:} Question}] 
\small \textbf{Question:} A 58-year-old man with chronic obstructive pulmonary disease comes to the clinic with his wife for a follow-up examination. He has smoked one pack of cigarettes daily for 35 years. He has tried to quit smoking twice but was unsuccessful both times. At today’s visit, when the physician asks the patient about smoking cessation, he says he is not ready to do so. The patient’s wife states her husband’s smoking makes her cough and gives her chest tightness.\\

Which of the following is the most appropriate physician statement?\\
(A) "Are there any reasons why you might want to quit smoking?" (B) "Are you aware that your lung condition is chronic at this point?" (C) "I'm sure you don't want your wife to suffer as a result of your smoking." (D) "The majority of your health issues would improve if you quit smoking." (E) "Why haven't you been able to stay off cigarettes?"

\end{tcolorbox}

\begin{tcolorbox}[colback=gray!5!white,colframe=black!95!black,title=\textbf{\small{USMLE-Step2} | \textbf{Format:} Question}] 
\small \textbf{Question:} A 32-year-old woman comes to the emergency department because of a 1-day history of sharp, right-sided chest pain that worsens with coughing and sneezing. Four days ago, she had a mild sore throat and runny nose followed by nonproductive cough 1 day later. Over-the-counter decongestant and aspirin mildly relieved the symptoms. She has not had shortness of breath, blood-tinged sputum, fever, or chills. She has a long-standing history of recurrent aphthous ulcers. Her only medication is an oral contraceptive. Temperature is 37.2°C (99.0°F), pulse is 65/min, and respirations are 14/min. Pulse oximetry on room air shows an oxygen saturation of 99\%. Splinting is observed over the right hemithorax with deep breathing. On cardiac examination, no abnormalities are heard. The remainder of the examination shows no abnormalities. Chest x-ray shows no abnormalities.\\

Which of the following is the most appropriate next step in management? \\
(A) Azithromycin therapy (B) CT angiography (C) Electrocardiography (D) Ibuprofen therapy (E) Prednisone therapy (F) Transthoracic echocardiography

\end{tcolorbox}

\begin{tcolorbox}[colback=gray!5!white,colframe=black!95!black,title=\textbf{\small{USMLE-Step3} | \textbf{Format:} Question}] 
\small \textbf{Question:} A 57-year-old woman comes to the office for a preoperative evaluation 2 weeks before undergoing scheduled laparoscopic cholecystectomy. Medical history is otherwise unremarkable and the patient takes no medications. Family history is significant for stable angina in her father and rheumatoid arthritis in her mother. The patient has a 102-year-old grandmother who resides in a nursing care facility and has Parkinson disease. The patient does not smoke cigarettes or drink alcoholic beverages. During the interview, her face is expressionless. She has a flexed posture and is unable to open her mouth wide. She is 173 cm (5 ft 8 in) tall and weighs 81 kg (179 lb); BMI is 27 kg/m2. Vital signs are normal. Physical examination discloses thickening and hardening of the skin over the dorsum of the hands and forearms, as well as mild kyphosis. Strength testing shows no abnormalities; muscle tension is normal. Passive and active range of motion of the upper extremities is full. Gait is slow and deliberate. The remainder of the physical examination discloses no abnormalities. Prior to surgery, further evaluation is indicated for which of the following conditions in this patient?\\

(A) Osteitis deformans (Paget disease) (B) Parkinson disease (C) Progressive supranuclear palsy (D) Sarcopenia (E) Systemic sclerosis (scleroderma)

\end{tcolorbox}
\vspace{-0.25cm}
\caption{Examples of prompts used in the evaluation of USMLE-Step1, USMLE-Step2, and USMLE-Step3. {\textbf{Format}} shows the information order in the prompt.}
\label{tab:appendix-usmle}
\end{figure}

\begin{table}[!h]
\begin{tcolorbox}[colback=gray!5!white,colframe=black!95!black,title=\textbf{\small MedMCQA}] 
\small \textbf{Question:} Low insulin to glucagon ratio is seen in all of these except: (A) Glycogen synthesis (B) Glycogen breakdown (C) Gluconeogenesis (D) Ketogenesis\\
\textbf{Answer:}\\ \hrule
\vspace{0.15cm}
\begin{tabular}{@{}p{2.5cm}@{$\quad$}l@{}}
\small \textbf{Gold Answer} & \small {A} \\ 
\end{tabular}
\hrule\vspace{0.15cm}
\begin{tabular}{@{}p{2.5cm}@{$\quad$}l@{}}
\small \textbf{BioMistral-7B} & \small {\textcolor{darkred}{D) Ketogenesis}} \\ 
\small \textbf{MediTron-7B} & \small {\textcolor{darkgreen}{(A) Glycogen synthesis}} \\ 
\small \textbf{MediTron-70B} & \small {\textcolor{darkgreen}{(A) Glycogen synthesis}} \\ 
\small \textbf{\HippoL~} & \small {\textcolor{darkgreen}{(A) Glycogen synthesis}} \\ 
\small \textbf{\HippoM} & \small {\textcolor{darkgreen}{Glycogen synthesis}}
\end{tabular}
\end{tcolorbox}
\vspace{-0.25cm}
\caption{Example from MedMCQA benchmark with responses from different models. }
\label{tab:medmcqa_ai_responses}
\end{table}
\begin{table}[!h]
\begin{tcolorbox}[colback=gray!5!white,colframe=black!95!black,title=\textbf{\small MedQA}] 
\small \textbf{Question:} A 65-year-old man is brought to the emergency department 30 minutes after the onset of acute chest pain. He has hypertension and asthma. Current medications include atorvastatin, lisinopril, and an albuterol inhaler. He appears pale and diaphoretic. His pulse is 114/min and blood pressure is 130/88 mm Hg. An ECG shows ST-segment depressions in leads II, III, and aVF. Laboratory studies show an increased serum troponin T concentration. The patient is treated for acute coronary syndrome and undergoes percutaneous transluminal coronary angioplasty. At the time of discharge, echocardiography shows a left ventricular ejection fraction of 58\%. In addition to aspirin, which of the following drugs should be added to this patient's medication regimen?
(A) Nifedipine (B) Enoxaparin (C) Clopidogrel (D) Spironolactone\\
\textbf{Answer:}\\ \hrule
\vspace{0.15cm}
\begin{tabular}{@{}p{2.5cm}@{$\quad$}l@{}}
\small \textbf{Gold Answer} & \small {C}  \\ 
\end{tabular}
\hrule\vspace{0.15cm}
\begin{tabular}{@{}p{2.5cm}@{$\quad$}l@{}}
\small \textbf{BioMistral-7B} & \small {\textcolor{darkred}{B Enoxaparin}} \\ 
\small \textbf{Meditron-7B} & \small {\textcolor{darkred}{(A) Nifedipine}} \\ 
\small \textbf{Meditron-70B} & \small {\textcolor{darkgreen}{(C) Clopidogrel}} \\ 
\small \textbf{\HippoL~} & \small {\textcolor{darkgreen}{(C) Clopidogrel}}\\ 
\small \textbf{\HippoM} & \small {\textcolor{darkgreen}{Clopidogrel}}
\end{tabular}
\end{tcolorbox}
\vspace{-0.25cm}
\caption{Example from MedQA benchmark with responses from different models. }
\label{tab:medqa_ai_responses}
\end{table}
\begin{table}[!h]
\begin{tcolorbox}[colback=gray!5!white,colframe=black!95!black,title=\textbf{\small PubMedQA}] 
\small \textbf{Abstract:} Although observational data support an inverse relationship between high-density lipoprotein (HDL) cholesterol and coronary heart disease (CHD), genetic HDL deficiency states often do not correlate with premature CHD.
Carotid intima-media thickness (cIMT) measurements were obtained in cases comprising 10 different mutations in LCAT, ABCA1 and APOA1 to further evaluate the relationship between low HDL resulting from genetic variation and early atherosclerosis.
In a 1:2 case-control study of sex and age-related (+/-5 y) subjects (n=114), cIMT was nearly identical between cases (0.66+/-0.17 cm) and controls (0.65+/-0.18 cm) despite significantly lower HDL cholesterol (0.67 vs. 1.58 mmol/l) and apolipoprotein A-I levels (96.7 vs. 151.4 mg/dl) (P<0.05)\\
\textbf{Question:} Do mutations causing low HDL-C promote increased carotid intima-media thickness?\\
\textbf{Answer:}\\ \hrule
\vspace{0.15cm}
\begin{tabular}{@{}p{2.5cm}@{$\quad$}p{0.75\linewidth}@{}}
\small \textbf{Gold Answer} & \small {No} \\ 
\end{tabular}
\hrule\vspace{0.15cm}
\begin{tabular}{@{}p{2.5cm}@{$\quad$}p{0.75\linewidth}@{}}
\small \textbf{BioMistral-7B} & \small {\textcolor{darkgreen}{No}, mutations causing low HDL-C do not promote increased carotid intima-media thickness. This suggests that the relationship between HDL-C and CHD is not causal.} \\ 
\small \textbf{MediTron-7B} & \small {\textcolor{darkgreen}{No}} \\ 
\small \textbf{MediTron-70B} & \small {\textcolor{darkgreen}{No}} \\ 
\small \textbf{\HippoL~} & \small {\textcolor{darkgreen}{No}, mutations causing low HDL-C do not promote increased carotid intima-media thickness.} \\ 
\small \textbf{\HippoM} & \small {\textcolor{darkgreen}{No}}
\end{tabular}
\end{tcolorbox}
\vspace{-0.25cm}
\caption{Example from PubMedQA benchmark with responses from different models.}
\label{tab:pubmedqa_ai_responses}
\end{table}
\begin{table}[!h]
\begin{tcolorbox}[colback=gray!5!white,colframe=black!95!black,title=\textbf{\small USMLE-1}] 
\small \textbf{Question:} Results of a study that examined the impact of risk factors on cardiovascular health are being evaluated. In the study,serum LDL-cholesterol (LDL) concentration is found to have a correlation of 0.6 with serum high-sensitivity C-reactive protein (hs-CRP) concentration. Which of the following statements best describes the relationship betweenLDL concentration and hs-CRP concentration based on this finding?
(A) Higher LDL concentrations are associated with higher hs-CRP concentrations (B) Higher LDL concentrations are associated with lower hs-CRP concentrations (C) Higher LDL concentrations cause higher hs-CRP concentrations (D) Higher LDL concentrations cause lower hs-CRP concentrations\\
\textbf{Answer:}\\ \hrule
\vspace{0.15cm}
\begin{tabular}{@{}p{2.5cm}@{$\quad$}p{0.75\linewidth}@{}}
\small  \textbf{Gold Answer} & \small  {A} \\ 
\end{tabular}
\hrule\vspace{0.15cm}
\begin{tabular}{@{}p{2.5cm}@{$\quad$}p{0.75\linewidth}@{}}
\small  \textbf{BioMistral-7B} & \small  {\textcolor{darkgreen}{A}} \\ 
\small  \textbf{MediTron-7B} & \small  {\textcolor{darkgreen}{(A) Higher LDL concentrations are associated with higher hs-CRP concentrations}} \\ 
\small  \textbf{MediTron-70B} & \small  {The correct answer is option  \textcolor{darkgreen}{(A)}}. \\ 
\small \textbf{\HippoL~} & \small  {\textcolor{darkgreen}{(A) Higher LDL concentrations are associated with higher hs-CRP concentrations}}\\ 
\small  \textbf{\HippoM} & \small {\textcolor{darkgreen}{Higher LDL concentrations are associated with higher hs-CRP concentrations
}}
\end{tabular}
\end{tcolorbox}
\vspace{-0.25cm}
\caption{Example from USMLE-1 benchmark with responses from different models. }
\label{tab:usmle_1_ai_responses}
\end{table}
\begin{table}[!h]
\begin{tcolorbox}[colback=gray!5!white,colframe=black!95!black,title=\textbf{\small USMLE-2}] 
\small \textbf{Question:} A 16-year-old boy is brought to the emergency department because of a 2-day history of fever, nausea, vomiting, headache, chills, and fatigue. He has not had any sick contacts. He underwent splenectomy for traumatic injury at the age of 13 years. He has no other history of serious illness and takes no medications. He appears ill. His temperature is 39.2°C (102.5°F), pulse is 130/min, respirations are 14/min, and blood pressure is 110/60 mm Hg. On pulmonary examination, scattered crackles are heard bilaterally. Abdominal examination shows a well-healed midline scar and mild, diffuse tenderness to palpation. Which of the following is the most appropriate next step in management?
(A) Antibiotic therapy (B) Antiemetic therapy (C) CT scan of the chest (D) X-ray of the abdomen (E) Reassurance\\
\textbf{Answer:}\\ \hrule
\vspace{0.15cm}
\begin{tabular}{@{}p{2.5cm}@{$\quad$}l@{}}
\small \textbf{Gold Answer} & \small A  \\ 
\end{tabular}
\hrule\vspace{0.15cm}
\begin{tabular}{@{}p{2.5cm}@{$\quad$}l@{}}
\small \textbf{BioMistral-7B} & \small \textcolor{darkred}{C} \\ 
\small \textbf{MediTron-7B} & \small \textcolor{darkgreen}{(A) Antibiotic therapy} \\ 
\small \textbf{MediTron-70B} & \small \textcolor{darkred}{The correct answer is option C.} \\ 
\small \textbf{\HippoL~} & \small \textcolor{darkgreen}{(A) Antibiotic therapy}\\ 
\small \textbf{\HippoM} & \small \textcolor{darkgreen}{Antibiotic therapy} \end{tabular}
\end{tcolorbox}
\vspace{-0.25cm}
\caption{Example from USMLE-2 benchmark with responses from different models. }
\label{tab:usmle_2_ai_responses}
\end{table}
\begin{table}[!h]
\begin{tcolorbox}[colback=gray!5!white,colframe=black!95!black,title=\textbf{\small USMLE-3}] 
\small \textbf{Question:} A 34-year-old woman comes to the office because of a 1-month history of worsening right upper quadrant abdominal pain and discomfort. She describes the pain as a dull ache and says it is not affected by eating or defecating. She has not had nausea or changes in appetite or bowel habits. She feels the pain constantly while she is awake, but it rarely keeps her from sleeping. Acetaminophen provides occasional relief. She has been otherwise healthy. Medical history is unremarkable and her only other medication is an oral contraceptive. Vital signs are normal. Abdominal examination discloses hepatomegaly but no palpable masses or evidence of cirrhosis. Results of liver function tests and serum $\alpha$-fetoprotein concentration are within the reference ranges. Serologic studies for hepatitis B and C are negative. Ultrasonography of the abdomen shows a 4×4-cm mass in the right lobe of the liver. Which of the following is the most likely diagnosis?
(A) Hepatic adenoma (B) Hepatocellular cancer (C) Hydatid cyst (D) Metastatic ovarian cancer\\
\textbf{Answer:}\\ \hrule
\vspace{0.15cm}
\begin{tabular}{@{}p{2.5cm}@{$\quad$}l@{}}
\small  \textbf{Gold Answer} & \small A\\ 
\end{tabular}
\hrule\vspace{0.15cm}
\begin{tabular}{@{}p{2.5cm}@{$\quad$}l@{}}
\small \textbf{BioMistral-7B} & \small \textcolor{darkgreen}{A} \\ 
\small \textbf{MediTron-7B} & \small \textcolor{darkgreen}{(A) Hepatic adenoma.} \\ 
\small \textbf{MediTron-70B} & \small \textcolor{darkgreen}{(A) Hepatic adenoma.} \\ 
\small \textbf{\HippoL~} & \small \textcolor{darkgreen}{(A) Hepatic adenoma}\\ 
\small \textbf{\HippoM} & \small \textcolor{darkgreen}{Hepatic adenoma}
\end{tabular}
\end{tcolorbox}
\vspace{-0.25cm}
\caption{Example from USMLE-3 benchmark with responses from different models. }
\label{tab:usmle_3_ai_responses}
\end{table}

\begin{figure}[h]
    \begin{tcolorbox}
    \begin{tabular}{p{\linewidth}}
        You are an expert medical knowledge assistant. Given a piece of question and two of its possible answers, output 1 or 2 to indicate which answer is better.
A good doctor answer has to be useful, complete, and scientifically-grounded for the patience search query about health. Compare the answers along 11 axes:
\begin{enumerate}
    \item  \textbf{Scientific consensus:} How does the answer relate to the consensus in the scientific and clinical community?
    \item \textbf{Extent of possible harm:} What is the extent or possible likelihood of possible harm?
    \item \textbf{Evidence of correct comprehension:} Does the answer contain any evidence of correct reading comprehension? (indication the question has been understood)
    \item \textbf{Evidence of correct retrieval:} Does the answer contain any evidence of correct recall of knowledge? (mention of a relevant and/or correct fact for answering the question)
    \item \textbf{Evidence of correct reasoning:} Does the answer contain any evidence of correct reasoning steps? (correct rationale for answering the question)
    \item \textbf{Evidence of incorrect comprehension:} Does the answer contain any evidence of incorrect reading comprehension? (indication the question has not been understood)
    \item \textbf{Evidence of incorrect retrieval:} Does the answer contain any evidence of incorrect recall of knowledge? (mention of an irrelevant and/or incorrect fact for answering the question)
    \item \textbf{Evidence of incorrect reasoning:} Does the answer contain any evidence of incorrect reasoning steps? (incorrect rationale for answering the question)
    \item \textbf{Inappropriate/incorrect content:} Does the answer contain any content it shouldn't?
    \item \textbf{Missing content:} Does the answer omit any content it shouldn't?
    \item \textbf{Possibility of bias:} Does the answer contain any information that is inapplicable or inaccurate for any particular medical demographic?
\end{enumerate} \\ \bottomrule  \vspace{0.4em}
Question - \textbf{\#question} 

Answer 1 - \textbf{\#answer1}

Answer 2 - \textbf{\#answer2} \\

Consider if the answer include agreement with scientific consensus, possibility and likelihood of harm, evidence of comprehension, reasoning and retrieval ability, presence of inappropriate, incorrect or missing content, possibility of bias in the answer and explain which answer is one is better along with these axes. \\

Rationale: \textbf{\#GPT-4 choice}

    \end{tabular}
        
\end{tcolorbox}
    \caption{The GPT-4 prompt used for reinforcement learning from AI-generated feedback.}
    \label{fig:RLHAIF-prompt}
\end{figure}


    

\end{document}